\documentclass{article}

\usepackage{times}
\usepackage{epsfig}
\usepackage{graphicx}
\usepackage{amsmath}
\usepackage{amssymb}
\usepackage{wrapfig}
\usepackage{subcaption}
\usepackage{caption}

\usepackage{multirow}

\usepackage{array}
\newcommand{\PreserveBackslash}[1]{\let\temp=\\#1\let\\=\temp}
\newcolumntype{C}[1]{>{\PreserveBackslash\centering}p{#1}}
\newcolumntype{R}[1]{>{\PreserveBackslash\raggedleft}p{#1}}
\newcolumntype{L}[1]{>{\PreserveBackslash\raggedright}p{#1}}

\usepackage{booktabs} 
\usepackage{balance}

\newcommand\blfootnote[1]{%
  \begingroup
  \renewcommand\thefootnote{}\footnote{#1}%
  \addtocounter{footnote}{-1}%
  \endgroup
}

\usepackage{hyperref}
\usepackage{xcolor}

\usepackage[accepted]{icml2020}

\newcommand{\beginsupplement}{%
        \setcounter{table}{0}
        \renewcommand{\thetable}{A\arabic{table}}%
        \setcounter{figure}{0}
        \renewcommand{\thefigure}{A\arabic{figure}}%
        \setcounter{section}{0}
        \renewcommand{\thesubsection}{\Alph{subsection}}
        \renewcommand{\figurename}{Appendix, Figure}
     }

\icmltitlerunning{Few-shot Image Classification: Just Use a Library of Pre-trained Feature Extractors and a Simple Classifier}

\begin{document}

\twocolumn[
\icmltitle{Few-shot Image Classification: Just Use a Library of Pre-trained Feature Extractors and a Simple Classifier}




\begin{icmlauthorlist}
\icmlauthor{Arkabandhu Chowdhury}{rice}
\icmlauthor{Mingchao Jiang}{rice}
\icmlauthor{Swarat Chaudhuri}{uta}
\icmlauthor{Chris Jermaine}{rice}
\end{icmlauthorlist}

\icmlaffiliation{rice}{Rice University}
\icmlaffiliation{uta}{University of Texas, Austin}

\icmlkeywords{One-shot, Transfer learning, Meta-learning, Library classifier}

\vskip 0.3in
]



\printAffiliationsAndNotice{} 

\blfootnote{Accepted at ICCV 2021}

\begin{abstract}
Recent papers have suggested that transfer learning can outperform sophisticated meta-learning methods for few-shot image classification. We take this hypothesis to its logical conclusion, and suggest the use of an ensemble of high-quality, pre-trained feature extractors for few-shot image classification. We show experimentally that a library of pre-trained feature extractors combined with a simple feed-forward network learned with an L2-regularizer can be an excellent option for solving cross-domain few-shot image classification. Our experimental results suggest that this simple approach far outperforms several well-established meta-learning algorithms.

\end{abstract}

\section{Introduction}\label{sec:intro}

There has been a lot of recent interest in few-shot image classification \cite{fei2006one, lee2019meta, li2019finding, finn2017model, finn2018probabilistic, kim2018bayesian, nichol2018reptile, lee2018gradient}.  Various papers have explored different formulations of the problem, but in one general formulation, we are given a data set $\mathcal{D}_\mathit{trn}$ of (image, label) pairs sampled from a distribution $\mathbb{P}_\mathit{trn}$. The goal is to devise a method that uses $\mathcal{D}_\mathit{trn}$ to learn a function $f$ that is \emph{itself} a few-shot learner. The few-shot learner $f$ takes as input a new labeled data set $\mathcal{D}_\mathit{few}$ consisting of a set of samples from a new distribution $\mathbb{P}_\mathit{few} \neq \mathbb{P}_\mathit{trn}$. $f$ then returns a classification function $g$, which is targeted at classifying samples from the distribution $\mathbb{P}_\mathit{few}$.  

The process of learning $f$ is often referred to as \emph{meta-learning} in the literature.  Learning to classify samples from $\mathbb{P}_\mathit{few}$ is a ``few shot'' problem when $\mathcal{D}_\mathit{few}$ is small, perhaps having only one example for each class produced by $\mathbb{P}_\mathit{few}$.  In the most difficult and generally applicable variant of the few-shot problem---which we consider in this paper---$\mathbb{P}_\mathit{few}$ has no known relationship to $\mathbb{P}_\mathit{trn}$ (this is ``cross-domain'' few-shot learning) and $\mathcal{D}_\mathit{few}$ is not available while learning $f$.  Thus, the meta-learning process has no access to information about the eventual application. The only information we have about $\mathbb{P}_\mathit{few}$ is the set $\mathcal{D}_\mathit{few}$, and this is available only when constructing $g$.   We cannot, for example, choose any hyperparameters controlling the learning of $g$ using information not  extracted from $\mathcal{D}_\mathit{few}$. 

Our goal is to devise a learner $f$ that works well, out-of-the-box, on virtually any new distribution $\mathbb{P}_\mathit{few}$.  We argue that in such a scenario, developing novel meta-learning methods to learn $f$ from scratch on given $\mathcal{D}_\mathit{trn}$ may not be the most productive direction of inquiry. Because no information about $\mathbb{P}_\mathit{few}$ is available during meta-learning, it makes sense to choose a $\mathcal{D}_\mathit{trn}$ that has many different types of images, so it is likely to contain some images with features similar to those produced by $\mathbb{P}_\mathit{few}$, whatever form this distribution takes.  Fortunately, in computer image classification, the standard benchmark data set is now ILSVRC2012, a 1000-class version of the full ImageNet \cite{ILSVRC15}.  ILSVRC2012 consists of a wide variety of images, and it has become quite standard for researchers who design and train new image classifiers to publish classifiers trained on ILSVRC2012.  Such published artifacts represent thousands of hours of work by researchers who are well-versed in the ``black art'' of wringing every last percent of accuracy out of a deep CNN.  Instead of developing new meta-learning methods, it may be more productive to simply fix $\mathcal{D}_\mathit{trn}$ = ILSVRC2012, and attempt to leverage all of the effort that has gone into learning deep CNNs over ILSVRC2012, by using those CNNs as the basis for the few-shot learner $f$.  As other, even more wide-ranging and difficult benchmark data sets become prevalent (such as the full, 20,000+ class ImageNet), high-quality classifiers trained using that data set may be preferred.

We first show that it is possible to use any of a  number of published, high-quality, deep CNNs, learned over ILSVRC2012, as the basis for a few-shot learner that significantly outperforms state-of-the art methods.  The way to do this is embarrassingly simple: remove the classifier from the top of the deep CNN, fix the weights of the remaining deep feature extractor, and replace the classifier with a simple MLP that is trained using $L_2$ regularization to prevent over-fitting.  We call these ``library-based'' learners because they are based on standard, published feature extractors. 

Next, we ask: if a published deep CNN can be used to produce a state-of-the-art, few-shot learner, can we produce an even higher-quality few-shot learner by combining together \emph{many} high-quality, deep CNNs?   We call such a learner a ``full library'' learner.

Then, we note that other researchers have suggested the utility of re-using high-quality, pre-trained feature extractors for few-shot image classification.  In particular, the authors of the  ``Big Transfer'' paper \cite{kolesnikov2019big} argue that a very large network trained on a huge data set (the JFT-300M data set \cite{sun2017revisiting}, with 300 million images) can power an exceptionally accurate transfer-learning-based few-shot learning. Unfortunately, the authors have not made their largest JFT-300M-trained network public, and so we cannot experiment with it directly.  However, they \emph{have} made public several versions of their ``Big Transfer'' network, trained on the full, 20,000+ class ImageNet benchmark public.  Interestingly, we find that ``big'' may not be as important as ``diverse'': a single, few-shot classifier comprised of many different high-quality ILSVRC2012-trained deep CNNs seems to be a better option than a single few-shot classifier built on top of any of the Google-trained CNNs.
Finally, we investigate why a full library learner works so well.  We postulate two reasons for this.  First, having a very large number of features ($>$ 10,000) does not seem to be a problem for few-shot learning. Second, there seems to be strength in diversity, in the sense that different CNNs appear useful for different tasks.

\section{High Accuracy of Library-Based Learners}

\subsection{Designing a Library-Based Learner}

We begin by asking: what if we eschew advanced meta-learning methods, and instead simply use a very high-quality deep CNN pulled from a library, trained on the ILSVRC2012 data set, as the basis for a few-shot classifier? 

\begin{table*}[t]
\centering
\begin{small}

\begin{tabular}{l|cccccccr}
\toprule
     \multicolumn{1}{c}{} & Aircraft & FC100 & Omniglot & Texture & Traffic & Fungi & Quick Draw & \multicolumn{1}{c}{VGG Flower} \\ 
\midrule
     \multicolumn{9}{c}{5-way, 1-shot} \\
\midrule
\multirow{2}{*}{Worst}& 40.9 $\pm$ 0.9 & 50.8 $\pm$ 0.9 & 77.2$\pm$0.9 & 59.1 $\pm$ 0.9 & 55.5 $\pm$ 0.8 & 53.0 $\pm$ 0.9 & 57.3 $\pm$ 0.9 & 79.7 $\pm$ 0.8 \\
     & RN18 & DN121 & RN152 & DN169 & RN152 & DN201 & RN101 & RN18 \\
\hline
\multirow{2}{*}{Best}&  46.2 $\pm$ 1.0 & 61.2$\pm$0.9 & 86.5 $\pm$ 0.7 & 65.1 $\pm$ 0.9 & 66.6 $\pm$ 0.9 & 56.6 $\pm$ 0.9 & 62.8 $\pm$ 0.9 &  83.5 $\pm$ 0.8 \\
     & DN161 & RN152 & DN121 & RN101 & DN201 & DN121 & RN18 &  DN161 \\

\midrule
     \multicolumn{9}{c}{20-way, 1-shot} \\
\midrule 
\multirow{2}{*}{Worst}& 20.1 $\pm$ 0.3 & 27.8 $\pm$ 0.4 & 56.2 $\pm$ 0.5 & 38.0 $\pm$ 0.4 & 29.7 $\pm$ 0.3 &  31.7 $\pm$ 0.4 & 33.2 $\pm$ 0.5 & 62.4 $\pm$ 0.5 \\
     & RN101 & DN121 & RN101 & RN18 & RN101 & RN101 & RN101 &  RN101 \\
\hline
\multirow{2}{*}{Best} & 24.3 $\pm$ 0.3 & 36.4 $\pm$ 0.4 & 69.1 $\pm$ 0.5 & 42.5 $\pm$ 0.4 & 38.5 $\pm$ 0.4 & 33.9 $\pm$ 0.5 & 39.5$\pm$ 0.5 & 70.0 $\pm$ 0.5  \\
     & DN161 & RN152 & DN121 & RN152 & DN201 & DN161 & DN201 & DN161 \\
\midrule
     \multicolumn{9}{c}{40-way, 1-shot} \\
\midrule
\multirow{2}{*}{Worst}& 14.2 $\pm$ 0.2 & 19.6 $\pm$ 0.2 & 47.3 $\pm$ 0.3  & 28.9 $\pm$ 0.2  & 22.2 $\pm$ 0.2 & 23.7 $\pm$ 0.3  & 26.4 $\pm$ 0.3 & 53.1 $\pm$ 0.3  \\
     & RN34 & RN18 & RN152 & RN18 & RN152 & RN34 & RN152 &  RN34 \\
\hline
\multirow{2}{*}{Best}& 17.4 $\pm$ 0.2 & 27.2 $\pm$ 0.3 & 61.6 $\pm$ 0.3 & 33.2 $\pm$ 0.3 & 29.5 $\pm$ 0.2 & 26.8 $\pm$ 0.3 & 31.2 $\pm$ 0.3 & 62.8 $\pm$ 0.3  \\

     & DN161 & RN152 & DN201 & DN152 & DN201 & DN161 & DN201 & DN161  \\
\bottomrule

\end{tabular}
\end{small}
\vspace{5 pt}
\caption{Accuracy obtained using library deep CNNs for few-shot learning.}
\label{tab:library}
\end{table*}

Specifically, we are given a high-quality, pre-trained deep CNN, from a library of pre-trained networks; we take the CNN as-is, but remove the topmost layers used for classification. This results in a function that takes an image, and returns an embedding. We then use that embedding to build a classifier in an elementary fashion: we feed the embedding into a multi-layer perceptron with a single hidden layer; a softmax is  used to produce the final classification. Given a few-shot classification problem, two weight matrices $\textbf{W}_1$ and  $\textbf{W}_2$ are learned; the first connecting the embedding to the hidden layer, the second connecting the hidden layer to the softmax.  To prevent over-fitting during training, simple $L_2$ regularization is used on the weight matrices.

\subsection{Evaluation}

To evaluate this very simple few-shot learner, we first identify nine, high-quality deep CNNs with published models, trained on ILSVRC2012: ResNet18, ResNet34, ResNet50, ResNet101, ResNet152 (all of the ResNet implementations are the ones from the original ResNet designers \cite{he2016deep}), DenseNet121, DenseNet161, DenseNet169, and DenseNet201 (all of the DenseNet implementations are also from the original designers \cite{huang2017densely}).

Our goal is to produce an ``out-of-the-box'' few-shot learner that can be used on any (very small) training set $\mathcal{D}_\mathit{few}$ without additional data or knowledge of the underlying distribution. We are very careful not to allow validation or parameter tuning on testing data domains, so all parameters and settings need to be chosen apriori.  If it is possible to build such a few-shot learner, it would be the most widely applicable: simply produce a few training images for each class, apply the learner. Thus, we perform an extensive hyper-parameter search, solely using the Caltech-UCSD Birds 200 set \cite{WelinderEtal2010} as a validation data set, and then use the best hyperparameters from that data set in all of our experiments. Hyperparameters considered were learning rate, number of training epochs, regularization penalty weight, the number of neurons in the MLP hidden layer, and whether to drop the hidden layer altogether.  A separate hyper-parameter search was used for 5-way, 20-way, and 40-way classification.  

We then test the resulting few-shot learner---one learner per deep CNN---on eight different data sets, FGVC-Aircraft \cite{maji13fine-grained}, FC100 \cite{oreshkin2018tadam}, Omniglot \cite{lake2015human}, Traffic Sign \cite{houben2013detection}, FGCVx Fungi \cite{schroeder2018fgvcx}, Quick Draw \cite{jongejan2016quick}, and VGG Flower \cite{nilsback2008automated}.  To evaluate a few-shot learner on a data set, for an ``$m$-way $n$-shot'' classification problem, we randomly select $m$ different classes, and then randomly select $n$ images from each class for training; the remainder are used for testing.  As this is ``out-of-the-box'' learning, no validation is allowed.  

We performed this evaluation for $m$ in $\{5, 20, 40\}$ and $n$ in $\{1, 5\}$.  Due to space constraints, the full results are presented as supplementary material, and we give only a synopsis here (Table \ref{tab:comparison-5-5} and Table \ref{tab:comparison-20-5}).  In Table \ref{tab:library}, we show the best and worst accuracy achieved across the 9 learners, for each of the 8 data sets, for $m$ in $\{5, 20, 40\}$ and $n = 1$.

\begin{table*}[t]
\centering
\begin{small}
\begin{tabular}{l|cccccccr} 

\toprule
     \multicolumn{1}{c}{} & Aircraft & FC100 & Omniglot & Texture & Traffic & Fungi & Quick Draw & \multicolumn{1}{c}{VGG Flower} \\ 

\midrule
     Baseline & 47.6 $\pm$ 0.7 & 66.9 $\pm$ 0.7 & 96.5 $\pm$ 0.2 & 62.5 $\pm$ 0.7 & 82.1 $\pm$ 0.7 & 57.6 $\pm$ 1.7 & 75.6 $\pm$ 0.7 & 90.9 $\pm$ 0.5 \\

     Baseline++ & 40.9 $\pm$ 0.7 & 59.8 $\pm$ 0.8 & 90.6 $\pm$ 0.4 & 59.9 $\pm$ 0.7 & 79.6 $\pm$ 0.8 & 54.1 $\pm$ 1.7 & 68.4 $\pm$ 0.7 & 83.1 $\pm$ 0.7\\

     MAML & 33.1 $\pm$ 0.6 & 62.0 $\pm$ 0.8 & 82.6 $\pm$ 0.7 & 56.9 $\pm$ 0.8 & 67.4 $\pm$ 0.9 & 48.3 $\pm$ 1.8 & 77.5 $\pm$ 0.8 & 78.0 $\pm$ 0.7\\

     MatchingNet & 33.5 $\pm$ 0.6 & 59.4 $\pm$ 0.8 & 89.7 $\pm$ 0.5 & 54.7 $\pm$ 0.7 & 73.7 $\pm$ 0.8 & 55.7 $\pm$ 1.7 & 70.4 $\pm$ 0.8 & 74.2 $\pm$ 0.8\\

     ProtoNet & 41.5 $\pm$ 0.7 & 64.7 $\pm$ 0.8 & 95.5 $\pm$ 0.3 & 62.9 $\pm$ 0.7 & 75.0 $\pm$ 0.8 & 53.1 $\pm$ 1.8 & 74.9 $\pm$ 0.7 & 86.7 $\pm$ 0.6\\

     RelationNet & 37.5 $\pm$ 0.7 & 64.8 $\pm$ 0.8 & 91.2 $\pm$ 0.4 & 60.0 $\pm$ 0.7 & 68.6 $\pm$ 0.8 & 58.7 $\pm$ 1.8 & 71.9 $\pm$ 0.7 & 80.6 $\pm$ 0.7\\

     Meta-transfer & 46.2 $\pm$ 0.7  & 75.7 $\pm$ 0.8 & 93.5 $\pm$ 0.4 & 70.5 $\pm$ 0.7 & 80.0 $\pm$ 0.8 & 66.1 $\pm$ 0.8 & 77.7 $\pm$ 0.7 & 90.5 $\pm$ 0.6 \\
     
     FEAT & 46.2 $\pm$ 0.9  & 60.5 $\pm$ 0.8 & 85.3 $\pm$ 0.7 & 70.7 $\pm$ 0.7 & 70.5 $\pm$ 0.8 & 67.3 $\pm$ 1.7 & 69.9 $\pm$ 0.7 & 92.1 $\pm$ 0.4 \\
     
     SUR & 45.2 $\pm$ 0.8  & 67.2 $\pm$ 1.0 & \textbf{98.7 $\pm$ 0.1} & 59.6 $\pm$ 0.7 & 70.6 $\pm$ 0.8 & 60.0 $\pm$ 1.8 & 73.5 $\pm$ 0.7 & 90.8 $\pm$ 0.5 \\

\midrule
     Worst library- & 61.0 $\pm$ 0.9 & 71.9 $\pm$ 0.8& 94.0 $\pm$ 0.4& 79.3 $\pm$ 0.6 & 78.8 $\pm$ 0.7 & 77.1 $\pm$ 0.8 & 77.8 $\pm$ 0.7 & 95.3 $\pm$ 0.4 \\
     based & RN34 & DN121 & RN152 & RN18 & RN152 & RN34 & RN152 & RN34 \\

     Best library- & \textbf{66.0 $\pm$ 0.9} & \textbf{80.0 $\pm$ 0.6} & 96.7 $\pm$ 0.2 & \textbf{83.4 $\pm$ 0.6} & \textbf{85.3 $\pm$ 0.7} & \textbf{79.1 $\pm$ 0.7} & \textbf{81.8 $\pm$ 0.6} & \textbf{96.8 $\pm$ 0.3} \\
     based & DN161 & RN152 & DN201 & DN161 & DN201 & DN121 & DN201 & DN161 \\
\bottomrule

\end{tabular}
\end{small}
\vspace{5 pt}
\caption{Comparing competitive methods with the simple library-based learners, on the 5-way, 5-shot problem.}
\label{tab:comparison-5-5}
\end{table*}

To give the reader an idea of how this accuracy compares to the state-of-the-art, we compare these results with a number of few-shot learners from the literature.  We compare against Baseline and Baseline++ \cite{chen2019closer}, MAML \cite{finn2017model}, MatchingNet \cite{vinyals2016matching}, ProtoNet \cite{snell2017prototypical}, RelationNet \cite{sung2018learning}, Meta-transfer \cite{sun2019meta}, FEAT \cite{ye2020few}, and SUR \cite{2020arXiv200309338D}. When a deep CNN classifier must be chosen for any of these methods, we use a ResNet18. For methods that require a pre-trained CNN (FEAT, Meta-transfer, and SUR), we use the ResNet18 trained by the ResNet designers \cite{he2016deep}. Lest the reader be concerned that we chose the worst option (ResNet18), we point out that of the library-based few-shot learners, on the 5-way, 5-shot problem ResNet18 gave the best accuracy out of all of the ResNet-based learners for two of the data sets (see Table \ref{tab:complete-5-5}).  Further, these competitive methods tend to be quite expensive to train---MAML, for example, requires running  gradient descent over a gradient descent---and for such a method, the shallower ResNet18 is a much more reasonable choice than the deeper models (even using a ResNet18, we could not successfully train first-order MAML for 5-shot, 40-way classification, due to memory constraints).

For the competitive methods (other than SUR) we follow the same procedure as was used for the library-based few-shot classifiers: any training that is necessary is performed on the ILSVRC2012 data set, and hyperparameter validation is performed using the Caltech-UCSD Birds 200 data set.  Each method is then used without further tuning on the remaining eight data sets. To evaluate SUR on data set $X$, we use feature extractors trained on the data sets in \{Omniglot, Aircraft, Birds, Texture, Quickdraw, Flowers, Fungi,and ILSVRC\} $-$ $X$. Traffic Sign and FC100 datasets are reserved for testing only.

\begin{table*}[t]
\centering
\begin{small}
\begin{tabular}{l|cccccccr} 

\toprule
     \multicolumn{1}{c}{} & Aircraft & FC100 & Omniglot & Texture & Traffic & Fungi & Quick Draw & \multicolumn{1}{c}{VGG Flower} \\ 

\midrule
     Baseline & 24.2 $\pm$ 0.3 & 40.0 $\pm$ 0.4 & 87.5 $\pm$ 0.3 & 37.0 $\pm$ 0.3 & 59.9 $\pm$ 0.4 & 32.5 $\pm$ 0.8 & 52.8 $\pm$ 0.4 & 76.7 $\pm$ 0.4 \\

     Baseline++ & 18.4 $\pm$ 0.3 & 33.8 $\pm$ 0.3 & 76.2 $\pm$ 0.2 & 34.8 $\pm$ 0.3 & 55.3 $\pm$ 0.4 & 28.2 $\pm$ 0.8 & 45.5 $\pm$ 0.4 & 64.0 $\pm$ 0.4\\

     MAML & 11.8 $\pm$ 0.2 & 25.7 $\pm$ 0.3 & 46.5 $\pm$ 0.4 & 21.9 $\pm$ 0.3 & 27.0 $\pm$ 0.3 & 17.3 $\pm$ 0.7 & 30.7 $\pm$ 0.3 & 32.9 $\pm$ 0.3\\

     MatchingNet & 11.9 $\pm$ 0.2 & 31.6 $\pm$ 0.3 & 64.6 $\pm$ 0.6 & 31.6 $\pm$ 0.3 & 46.5 $\pm$ 0.4 & 28.1 $\pm$ 0.8 & 41.2 $\pm$ 0.4 & 53.7 $\pm$ 0.5\\

     ProtoNet & 22.1 $\pm$ 0.3 & 38.9 $\pm$ 0.4 & 88.1 $\pm$ 0.2 & 38.9 $\pm$ 0.3 & 46.9 $\pm$ 0.4 & 33.0 $\pm$ 0.9 & 33.0 $\pm$ 0.9 & 70.9 $\pm$ 0.4\\

     RelationNet & 17.1 $\pm$ 0.3 & 39.1 $\pm$ 0.4 & 79.7 $\pm$ 0.3 & 32.1 $\pm$ 0.3 & 41.9 $\pm$ 0.4 & 27.8 $\pm$ 0.8 & 47.5 $\pm$ 0.4 & 62.5 $\pm$ 0.4\\

     Meta-transfer & 19.1 $\pm$ 0.3  & 48.0 $\pm$ 0.4 & 75.3 $\pm$ 0.4 & 45.5 $\pm$ 0.4 & 52.0 $\pm$ 0.4 & 38.6 $\pm$ 0.5 & 52.6 $\pm$ 0.4 & 74.0 $\pm$ 0.4 \\
     
     FEAT & 23.1 $\pm$ 0.5  & 34.4 $\pm$ 0.5 & 66.4 $\pm$ 0.6 & 47.5 $\pm$ 0.6 & 43.4 $\pm$ 0.6 & 43.9 $\pm$ 0.8 & 47.0 $\pm$ 0.6 & 80.5 $\pm$ 0.5 \\
     
     SUR & 21.8 $\pm$ 0.3  & 42.9 $\pm$ 0.5 & \textbf{96.3 $\pm$ 0.1} & 35.5 $\pm$ 0.4 & 46.7 $\pm$ 0.4 & 34.4 $\pm$ 0.9 & 54.2 $\pm$ 0.4 & 77.1 $\pm$ 0.4 \\

\midrule
     Worst library- & 37.5 $\pm$ 0.4 & 47.1 $\pm$ 0.4 & 84.3 $\pm$ 0.3 & 58.7 $\pm$ 0.4 & 55.9 $\pm$ 0.4 & 56.1 $\pm$ 0.5 & 57.2 $\pm$ 0.4 & 86.8 $\pm$ 0.3 \\
     based & RN18 & RN18 & RN101 & RN18 & RN152 & RN34 & RN101 & RN101 \\

     Best library- & \textbf{44.6 $\pm$ 0.4} & \textbf{58.8 $\pm$ 0.4} & 92.0 $\pm$ 0.2 & \textbf{65.1 $\pm$ 0.4} & \textbf{66.0 $\pm$ 0.4} & \textbf{60.8 $\pm$ 0.5} & \textbf{63.9 $\pm$ 0.4} & \textbf{91.6 $\pm$ 0.2} \\
     based & DN161 & RN152 & DN201 & DN161 & DN201 & DN161 & DN161 & DN161 \\
\bottomrule

\end{tabular}
\end{small}
\vspace{5 pt}
\caption{Comparing competitive methods with the simple library-based learners, on the 20-way, 5-shot problem.}
\label{tab:comparison-20-5}
\end{table*}

A comparison of each of these competitive methods with the the best and worst-performing library-based learners on the 5-way, 5-shot learning problem is shown in Table \ref{tab:comparison-5-5}; a comparison on 20-way, 5-shot learning in Table \ref{tab:comparison-20-5}.  A more complete set of results is in the supplementary material.

\subsection{Discussion}

There are a few key takeaways from these results.  The best library-based learner \emph{always} beat every one of the other methods tested, with the only exception of SUR when testing on Omniglot data set. For the other data sets, the gap only grows as the number of ways increases. In fact, for the 20-way problem, the \emph{worst} library-based learner always beat all of the other methods tested (except SUR on Omniglot).  The gap can be quite dramatic, especially on the 20-way problem.  The best non-transfer based few-shot learner (MAML, Proto Nets, Relation Nets, and Matching Nets fall in this category) was far worse than even the worst library-based learner: 39\% accuracy for Proto Nets vs. 59\% accuracy for a classifier based on RestNet18 on Texture,  48\% accuracy for Relation Nets vs. 57\% accuracy for classifier based on ResNet101 on Quick Draw.  There have been a large number of non-transfer-based methods proposed in the literature (with a lot of focus on improving MAML in particular \cite{finn2017model, finn2018probabilistic, kim2018bayesian, nichol2018reptile, lee2018gradient}) but the gap between MAML and the library-based classifiers is very large.

We also note that of the rest non-library methods, Meta-transfer, Baseline, and FEAT were generally the best. We note that Meta-transfer, Baseline, and FEAT use the pre-trained ResNet18 without modification.  This tends to support the hypothesis at the heart of this paper: starting with a state-of-the-art feature extractor, trained by experts, may be the most important decision in few-shot learning.

\section{A Simple Full Library Classifier}

\subsection{Extreme Variation in Few-Shot Quality}

There is not a clear pattern to which of the library-based classifiers tends to perform best, and which tends to perform worst.  Consider the complete set of results, over all of the library-based few-shot learners, for the 5-way, 5-shot problem, shown in Table \ref{tab:complete-5-5}.  For ``out-of-the-box'' use, where no validation data are available, it is very difficult to see any sort of pattern that might help with picking a particular library-based classifier.  The DenseNet variations sometimes do better than ResNet (on the Aircraft data set, for example), but sometimes they do worse than the ResNet variations (on the FC100 data set, for example).  And within a family, it is unclear which library-based CNN to use.  As mentioned before, ResNet18 provides the best ResNet-based few-shot learner for two of the data sets, but it forms the basis of the worst ResNet-based learner on another two.

\begin{table*}[t]
\centering
\begin{small}
\begin{tabular}{l|cccccccr} 
\toprule
     \multicolumn{1}{c}{} & Aircraft & FC100 & Omniglot & Texture & Traffic & Fungi & Quick Draw & \multicolumn{1}{c}{VGG Flower} \\ 
\midrule 
     DenseNet121 & 64.7 $\pm$ 0.9 & 71.9 $\pm$ 0.8 & \textbf{96.7 $\pm$ 0.3} & 82.1 $\pm$ 0.6 & \textbf{85.0 $\pm$ 0.7} & \textbf{79.1 $\pm$ 0.7} & 81.3 $\pm$ 0.7 & 96.0 $\pm$ 0.4\\

     DenseNet161 & \textbf{66.0 $\pm$ 0.9} & 73.7 $\pm$ 0.7 & 96.6 $\pm$ 0.3 & \textbf{83.4 $\pm$ 0.6} & 83.9 $\pm$ 0.7& 78.4 $\pm$ 0.8 & 81.3 $\pm$ 0.6 & \textbf{96.8 $\pm$ 0.3}\\

     DenseNet169 & 63.6 $\pm$ 0.9 & 73.5 $\pm$ 0.7 & 95.0 $\pm$ 0.3 & 82.3 $\pm$ 0.6 & 84.6 $\pm$ 0.7 & 78.4 $\pm$ 0.8 & 80.6 $\pm$ 0.7 & 96.1 $\pm$ 0.3 \\

     DenseNet201 & 62.6 $\pm$ 0.9 & 75.1 $\pm$ 0.7 & \textbf{96.7 $\pm$ 0.6} & 83.2 $\pm$ 0.6 & 85.3 $\pm$ 0.7 & 78.0 $\pm$ 0.7 & \textbf{81.8 $\pm$ 0.6} & 96.5 $\pm$ 0.3 \\

     ResNet18 & 61.2 $\pm$ 0.9 & 72.1 $\pm$ 0.8 & 95.4 $\pm$ 0.3 & 79.3 $\pm$ 0.6 & 83.2 $\pm$ 0.7 & 77.7 $\pm$ 0.7 & 81.7 $\pm$ 0.6 & 95.3 $\pm$ 0.4 \\

     ResNet34 & 61.0 $\pm$ 0.9 & 76.2 $\pm$ 0.7 & 94.9 $\pm$ 0.3 & 82.5 $\pm$ 0.6 & 81.3 $\pm$ 0.7 & 77.1 $\pm$ 0.7 & 79.7 $\pm$ 0.6 & 95.3 $\pm$ 0.4 \\

     ResNet50 & 62.3 $\pm$ 0.9 & 73.9 $\pm$ 0.8 & 94.3 $\pm$ 0.4 & 83.2 $\pm$ 0.6 & 79.4 $\pm$ 0.8 & 77.9 $\pm$ 0.8 & 78.1 $\pm$ 0.8 & 95.6 $\pm$ 0.4 \\

     ResNet101 & 62.4 $\pm$ 0.9 & 79.2 $\pm$ 0.7 & 95.5 $\pm$ 0.3 & 82.8 $\pm$ 0.6 & 81.3 $\pm$ 0.7 & 78.6 $\pm$ 0.8 & 78.6 $\pm$ 0.7 & 95.8 $\pm$ 0.3\\

     ResNet152 & 61.7 $\pm$ 1.0 & \textbf{80.0 $\pm$ 0.6} & 94.0 $\pm$ 0.4 & 83.0 $\pm$ 0.6 & 78.8 $\pm$ 0.7 & 79.0 $\pm$ 0.8 & 77.8 $\pm$ 0.7 & 95.6 $\pm$ 0.3\\
\bottomrule

\end{tabular}
\end{small}
\vspace{5 pt}
\caption{Complete results for library-based few-shot learners on the  5-way, 5-shot problem.}
\label{tab:complete-5-5}
\end{table*}

\subsection{Combining Library-Based Learners}

Given the relatively high variance in accuracies obtained using the library deep CNNs across various data sets, it is natural to ask: Is it perhaps possible to use all of these library feature extractors in concert with one another, to develop a few-shot learner which consistently does as well as (or even better then) the best library CNN?

Given the lack of training data in few-shot learning, the first idea that one might consider is some simple variant of ensembling: given a few-shot learning problem, simply train a separate neural network on top of each of the deep CNNs, and then use a majority vote at classification time (hard ensembling) or we can average the class weights at classification time (soft ensembling). 

Another option is to take all of the library deep CNNs together, and view them as a single feature extractor.  Using the nine models considered thus far in this paper in this way results in 13,984 features. We then train an MLP on top of this, using $L_2$ regularization. Again using the Caltech-UCSD Birds 200 data set for validation, we perform hyperparameter search to build a few-shot learner for $5$-way, $20$-way, and $40$-way classification problems.

We test both of these options over the eight test sets, and give a synopsis of the results in Table \ref{tab:ensemble}. We find that across all 24 test cases (eight data sets, three classification tasks), the best single learner was \emph{never} able to beat the best method that used all nine deep CNNs.  All of the tested methods had similar performance on the 5-way classification task, though the best single learner was generally a bit worse than the  other methods.  

Where the difference becomes more obvious is on the classification tasks with more categories.  For the $20$-way and $40$-way problems, the two ensemble-based methods had a small but consistent drop in accuracy, and building a single network on top of all nine deep CNNs is clearly the best.  This may be somewhat surprising; for the 40-way problem, hyperparameter search on the Caltech-UCSD Birds 200 data set settled on a single network with 1024 neurons in a hidden layer; this means that more than 14 million parameters must be learned over just 200 images.  Clearly, the neural network is massively over-parameterized, and yet the accuracies obtained are remarkable, with over 90\% accuracy on two of the data sets.

\begin{table*}[t]
\centering
\begin{small}

\begin{tabular}{l|cccccccr} 
\toprule
     \multicolumn{1}{c}{} & Aircraft & FC100 & Omniglot & Texture & Traffic & Fungi & Quick Draw & \multicolumn{1}{c}{VGG Flower} \\ 
\midrule 
     \multicolumn{1}{c}{} & \multicolumn{8}{c}{5-way, 5-shot} \\
\midrule
     Full Library & \textbf{68.9 $\pm$ 0.9 } & 79.1 $\pm$ 0.8 & 97.5 $\pm$ 0.3 & 85.3 $\pm$ 0.6 & \textbf{85.8 $\pm$ 0.7} & 81.2 $\pm$ 0.8 & \textbf{84.2 $\pm$ 0.6} & 97.4 $\pm$ 0.3 \\
     Hard Ensemble& 67.8 $\pm$ 0.9 & 79.9 $\pm$ 0.7 & 97.8 $\pm$ 0.2 & 85.4 $\pm$ 0.5 & 85.2 $\pm$ 0.7 & \textbf{82.1 $\pm$ 0.7} & 83.5 $\pm$ 0.6 & 97.7 $\pm$ 0.2 \\
     Soft Ensemble& 68.4 $\pm$ 0.9 & \textbf{80.5 $\pm$ 0.6} & \textbf{98.0 $\pm$ 0.2} & \textbf{85.7 $\pm$ 0.6 } & 85.2 $\pm$ 0.7 & 82.0 $\pm$ 0.7 & 84.1 $\pm$ 0.5 &  \textbf{97.9 $\pm$ 0.2} \\
     Best Single & 66.0 $\pm$ 0.9 & 80.0 $\pm$ 0.6 & 96.7 $\pm$ 0.2 & 83.4 $\pm$ 0.6 & 85.2 $\pm$ 0.7 & 79.1 $\pm$ 0.7 & 81.8 $\pm$ 0.6 & 96.8 $\pm$ 0.3\\

\midrule
     \multicolumn{1}{c}{} & \multicolumn{8}{c}{20-way, 5-shot} \\
\midrule 
     Full Library & \textbf{49.5 $\pm$ 0.4} & \textbf{61.6 $\pm$ 0.4} & \textbf{95.4 $\pm$ 0.2} & \textbf{68.5 $\pm$ 0.4} & \textbf{70.4 $\pm$ 0.4} & \textbf{65.5 $\pm$ 0.5} & \textbf{69.4 $\pm$ 0.4} & \textbf{94.3 $\pm$ 0.2}  \\
     Hard Ensemble & 46.6 $\pm$ 0.4 & 60.1 $\pm$ 0.4 & 94.5 $\pm$ 0.2 & 67.8 $\pm$ 0.4 & 67.8 $\pm$ 0.4 & 64.4 $\pm$ 0.4  & 67.9 $\pm$ 0.4 & 93.5 $\pm$ 0.2  \\
     Soft Ensemble & 47.5 $\pm$ 0.4 & 60.7 $\pm$ 0.4 & 94.9 $\pm$ 0.2 & 68.2 $\pm$ 0.4 & 68.3 $\pm$ 0.4 & 64.4 $\pm$ 0.4 & 68.8 $\pm$ 0.4 & 93.7 $\pm$ 0.2 \\
     Best Single & 44.6 $\pm$ 0.4 & 58.8 $\pm$ 0.4 & 92.0 $\pm$ 0.2 & 65.1 $\pm$ 0.4 & 66.0 $\pm$ 0.4 & 60.8 $\pm$ 0.5 & 63.9 $\pm$ 0.4 & 91.6 $\pm$ 0.2\\
\midrule
     \multicolumn{1}{c}{} & \multicolumn{8}{c}{40-way, 5-shot} \\
\midrule 
     Full Library & \textbf{41.2 $\pm$ 0.3} & \textbf{51.8 $\pm$ 0.2} & \textbf{93.2 $\pm$ 0.1 } & \textbf{59.3 $\pm$ 0.2} & \textbf{62.7 $\pm$ 0.2} & \textbf{57.6 $\pm$ 0.3} & \textbf{60.8 $\pm$ 0.3} & \textbf{91.9 $\pm$ 0.2} \\
     Hard Ensemble& 38.0 $\pm$ 0.3 & 50.2 $\pm$ 0.2 & 92.1 $\pm$ 0.1 & 58.5 $\pm$ 0.2 & 59.6 $\pm$ 0.2 & 55.6 $\pm$ 0.3 & 59.3 $\pm$ 0.3 & 90.6 $\pm$ 0.2  \\
     Soft Ensemble & 39.0 $\pm$ 0.3 & 51.2 $\pm$ 0.3 & 92.5 $\pm$ 0.1 & 59.0 $\pm$ 0.2 & 60.2 $\pm$ 0.2 & 56.5 $\pm$ 0.3 & 60.1 $\pm$ 0.3 & 91.1 $\pm$ 0.2 \\
     Best Single & 35.9 $\pm$ 0.2 & 48.2 $\pm$ 0.3 & 89.4 $\pm$ 0.2 & 55.4 $\pm$ 0.2 & 57.5 $\pm$ 0.2 & 52.1 $\pm$ 0.3 & 55.5 $\pm$ 0.3 & 88.9 $\pm$ 0.2 \\
\bottomrule

\end{tabular}
\end{small}
\vspace{5 pt}
\caption{Accuracy obtained using all nine library CNNs as the basis for a few-shot learner.}
\label{tab:ensemble}
\end{table*}

\section{Data vs. Diversity: Who Wins?}

The results from the last section clearly show that an MLP learned on top of a library of high-quality, deep CNNs makes for an excellent, general-purpose, few-shot learner.  This leaves open the question.  When basing a few-shot learner on a fairly simple, transfer-based approach, which is more important, diversity or size, when constructing the few-shot learner?  That is, is it better to have a large variety of deep CNNs to combine together, each of which is trained on a smaller $\mathcal{D}_\mathit{trn}$, or is it preferable to base a few-shot learner on a single, deep CNN that has been trained on a larger and more diverse $\mathcal{D}_\mathit{trn}$?

To answer this question, we compare the single MLP built upon all nine of the library deep CNNs with an MLP built upon a high-quality deep CNN that was \emph{itself} constructed upon an even larger data set: the full ImageNet data set, with more than 20,000 categories.  High-quality, publicly available deep CNNs for this data set are rare, but Google recently released a set of deep CNNs trained on the full ImageNet, specifically for use in transfer learning\cite{kolesnikov2019big}. We consider three of their deep CNNs. Each is a ResNet: BiT-ResNet-101-3 (``BiT'' stands for ``Big Transfer''; ``101-3'' is a ResNet101 that has $3X$ the width of a standard ResNet), BiT-ResNet-152-4, and BiT-ResNet-50-1.  

For each of these Big Transfer models, we perform a full hyperparameter search using the Caltech-UCSD Birds data set for validation, as we did for each of the deep CNNs trained on ILSVRC2012.  Interestingly, the Google models tend to do better with a much larger $L_2$ regularization parameter weight (0.5 to 0.7) compared to the other deep CNNs trained on ILSVRC2012 (which typically performed best on the validation set using a weight of around 0.1). Results are shown in Table \ref{tab:vs-bigt}.

The headline finding is that the single model utilizing a library of nine, ILSVRC2012-trained CNNs, is the best model.  It did not not always perform the best on each data set. In fact, on four of the data sets (FC100, Texture, Fungi, and VGG Flower) at least one of the Google Big Transfer models outperformed the single model consisting of nine ILSVRC2012-trained CNNs.  

In each of the those cases, the performance was comparable across models, except, perhaps, for the VGG Flower data set, where the best Big Transfer models always obtained more than 99\% accuracy.  However, on the other data sets (Aircraft, Omniglot, Trafic Sign, and QuickDraw) the combined model far outperformed any of the Big Transfer models. The gap was often significant, and the combined model outperformed the best Big Transfer-based model by an average of more than 11\% for the 40-way task.

It was also interesting that while the Big Transfer models were generally better than the ILSVRC2012-trained library CNNs, this varied across data sets. On the Aircraft and Omniglot data sets, for example, even the best \emph{individual} ILSVRC2012-trained library CNNs outperformed the Big Transfer models.

All of this taken together seems to suggest that, when building a transfer-based few-shot learner, having a large library of deep CNNs is at least as important---and likely \emph{more} important---than having access to a CNN that was trained on a very large $\mathcal{D}_\mathit{trn}$.

\section{Why Does This Work?}

\subsection{Few-Shot Fine-Tuning Is Surprisingly Easy}

We turn our attention to asking: why does using a full library of pre-trained feature extractors seem to work so well?
One reason is that fine-tuning appears to be very easy with a library of pre-trained features.  Consider the following simple experiment, designed to test whether the number of training points has much effect on the learned model.  

We choose a large number of 40-way, problems, over all eight of our benchmark data sets, and train two classifiers for each.  Both classifiers use all of the 13,984 features provided by the nine library feature extractors.  However, the first classifier is trained as a one-shot classifier, and the second classifier is trained using all of the available data in the data set.
Our goal is to see whether the sets of learned weights have a strong correspondence across the two learners; if they do, it is evidence that the number of training points has a relatively small effect.  Note that in a neural network with a hidden layer consisting of hundreds or thousands of neurons, there are likely to be a large number of learned weights that are of equal quality; in fact, simply permuting the neurons in the hidden layer results in a model that is functionally identical, but that has very different weight matrices.  Thus, we do not use a hidden layer in either classifier, and instead use a softmax directly on top of an output layer that linearly combines the input features.  Both the one-shot and full-data classifiers were learned without regularization.  

\begin{table*}[t]
\centering
\begin{small}

\begin{tabular} {l|cccccccr}

\toprule
     \multicolumn{1}{c}{} & Aircraft & FC100 & Omniglot & Texture & Traffic & Fungi & QDraw & \multicolumn{1}{c}{Flower} \\
\midrule
     \multicolumn{1}{c}{} & \multicolumn{8}{c}{5-way, 5-shot} \\
\midrule
     Full Library & \textbf{68.9 $\pm$ 0.9 } & 79.1 $\pm$ 0.8 & \textbf{97.5 $\pm$ 0.3 } & 85.3 $\pm$ 0.6 & \textbf{85.8 $\pm$ 0.7} & 81.2 $\pm$ 0.8 & \textbf{84.2 $\pm$ 0.6} & 97.4 $\pm$ 0.3 \\
     BiT-ResNet-101-3& 54.0 $\pm$ 1.1 & 78.6 $\pm$ 0.8 & 82.5 $\pm$ 1.2 & 82.0 $\pm$ 0.9 & 69.2 $\pm$ 0.9 & 81.2 $\pm$ 1.2 & 63.7 $\pm$ 1.1 & 99.6 $\pm$ 0.1 \\
     BiT-ResNet-152-4& 59.5 $\pm$ 1.0 & \textbf{80.9 $\pm$ 0.7} & 94.2 $\pm$ 0.5 & \textbf{85.4 $\pm$ 0.6} & 73.3 $\pm$ 0.8 & \textbf{82.5 $\pm$ 0.9} & 74.8 $\pm$ 0.8 &  \textbf{99.7 $\pm$ 0.1} \\
     BiT-ResNet-50-1& 61.9 $\pm$ 1.2 & 79.0 $\pm$ 0.8 & 87.2 $\pm$ 1.1 & 84.2 $\pm$ 0.6 & 75.6 $\pm$ 1.0  & \textbf{82.5 $\pm$ 0.8} & 71.5 $\pm$ 0.8 & 99.3 $\pm$ 0.2 \\
\midrule
     \multicolumn{1}{c}{} & \multicolumn{8}{c}{20-way, 5-shot} \\
\midrule 
     Full Library & \textbf{49.5 $\pm$ 0.4} & 61.6 $\pm$ 0.4 & \textbf{95.4 $\pm$ 0.2} & 68.5 $\pm$ 0.4 & \textbf{70.4 $\pm$ 0.4} & 65.5 $\pm$ 0.5 & \textbf{69.4 $\pm$ 0.4} & 94.3 $\pm$ 0.2  \\
     BiT-ResNet-101-3& 35.8 $\pm$ 0.4 & 60.4 $\pm$ 0.4 & 87.8 $\pm$ 0.3 & 69.6 $\pm$ 0.4 & 51.1 $\pm$ 0.4 & 68.4 $\pm$ 0.5 & 57.0 $\pm$ 0.4 & 99.3 $\pm$ 0.1 \\
     BiT-ResNet-152-4& 33.5 $\pm$ 0.4 & \textbf{63.4 $\pm$ 0.4} & 85.4 $\pm$ 0.4 & \textbf{70.9 $\pm$ 0.4} & 49.2 $\pm$ 0.4 & 68.1 $\pm$ 0.5 & 52.6 $\pm$ 0.5 &  \textbf{99.5 $\pm$ 0.1} \\
     BiT-ResNet-50-1& 39.6 $\pm$ 0.4 & 60.9 $\pm$ 0.4 & 83.9 $\pm$ 0.4 & 66.4 $\pm$ 0.4 & 53.5 $\pm$ 0.4 & \textbf{68.7 $\pm$ 0.4} & 55.0 $\pm$ 0.4 &   99.1 $\pm$ 0.1 \\
\midrule
     \multicolumn{1}{c}{} & \multicolumn{8}{c}{40-way, 5-shot} \\
\midrule 
     Full Library & \textbf{41.2 $\pm$ 0.3} & 51.8 $\pm$ 0.2 & \textbf{93.2 $\pm$ 0.1} & 59.3 $\pm$ 0.2 & \textbf{62.7 $\pm$ 0.2} & 57.6 $\pm$ 0.3 & \textbf{60.8 $\pm$ 0.3} & 91.9 $\pm$ 0.2 \\
     BiT-ResNet-101-3& 24.6 $\pm$ 0.3 & 49.6 $\pm$ 0.2 & 56.4 $\pm$ 0.8 & \textbf{61.5 $\pm$ 0.2} & 40.2 $\pm$ 0.2 & \textbf{60.3 $\pm$ 0.3} & 28.9 $\pm$ 0.5 & 99.0 $\pm$ 0.1 \\
     BiT-ResNet-152-4& 25.4 $\pm$ 0.2 & \textbf{53.0 $\pm$ 0.3} & 81.0 $\pm$ 0.3 & 58.6 $\pm$ 0.2 & 40.0 $\pm$ 0.2 & 53.9 $\pm$ 0.4 & 44.8 $\pm$ 0.3 &  98.8 $\pm$ 0.1 \\
     BiT-ResNet-50-1& 33.0 $\pm$ 0.3 & 48.8 $\pm$ 0.3 & 84.6 $\pm$ 0.2 & 60.0 $\pm$ 0.2 & 46.9 $\pm$ 0.2 & 59.2 $\pm$ 0.3 & 48.3 $\pm$ 0.3 & \textbf{99.0 $\pm$ 0.1} \\
\bottomrule

\end{tabular}
\end{small}
\vspace{5 pt}
\caption{Comparing a few-shot learner utilizing the full library of nine ILSVRC2012-trained deep CNNs with the larger CNNs trained on the full ImageNet.}
\label{tab:vs-bigt}
\end{table*}

Over each of the eight benchmark data sets, for each feature, we take the $L_1$ norm of all of the 40 weights associated with the feature; this serves as an indication of the importance of the feature. We compute the Pearson correlation coefficient for each of the 13,984 norms obtained using the models learned for the 1-shot and full data classifiers.  These correlations are shown in Table \ref{tab:correlation}.

What we see is a remarkably high correlation between the two sets of learned weights; above 80\% correlation in every case, except for the Traffic Sign data set.  However, even in the case of the Traffic Sign data set, there was a weak correlation (the Traffic Sign data is a bit of an outlier in other ways as well, as we will see subsequently).

This would seem to indicate that there is a strong  signal with only one data point per class, as the learned weights do not differ much when they are learned using the whole data set. This may be one explanation for the surprising accuracy of the full library few-shot learner.  Of course, the strong correlation may gloss over significant differences, and more data \emph{does} make a significant difference (consider the difference between the one-shot accuracies in Table 1 and the five-shot accuracies in Table 2.).  But even one image per class seems to give a lot of information.

\begin{table*}[t]
\centering
\begin{small}

\begin{tabular}{l|ccccccccr}
\toprule
     \multicolumn{1}{c}{} & Aircraft & Birds & FC100 & Fungi & Omniglot & Quick Draw & Texture & Traffic  & \multicolumn{1}{c}{VGG Flower} \\ 
\midrule 
     Correlation & 0.95 & 0.88 & 0.89 & 0.86 & 0.93 & 0.92 & 0.80 & 0.18  & 0.87 \\
\bottomrule

\end{tabular}
\end{small}
\vspace{5 pt}
\caption{Correlation between weight learned using one example per class, and the full data.}
\label{tab:correlation}
\end{table*}

\subsection{Different Problems Utilize Different Features}

Another reason for the accuracy obtained by the full library method may be that different problem domains seem to utilize different sets of features, and different feature extractors.  Having a large library ensures that \emph{some} features relevant to any task are always present.

To investigate this, we construct a large number of 40-way training tasks over each of the various data sets, and for each, we learn a network without a hidden layer on top of all 13,984 features obtained using the library of deep CNNs. Again, we compute the $L_1$ norm of the set of weights associated with each of the features.  This time, however, for each problem we then consider the features whose norms are in the top 20\%; these could be considered the features most important to solving the classification task.

\begin{figure}{}
  \centering
    \includegraphics[width=.99\linewidth]{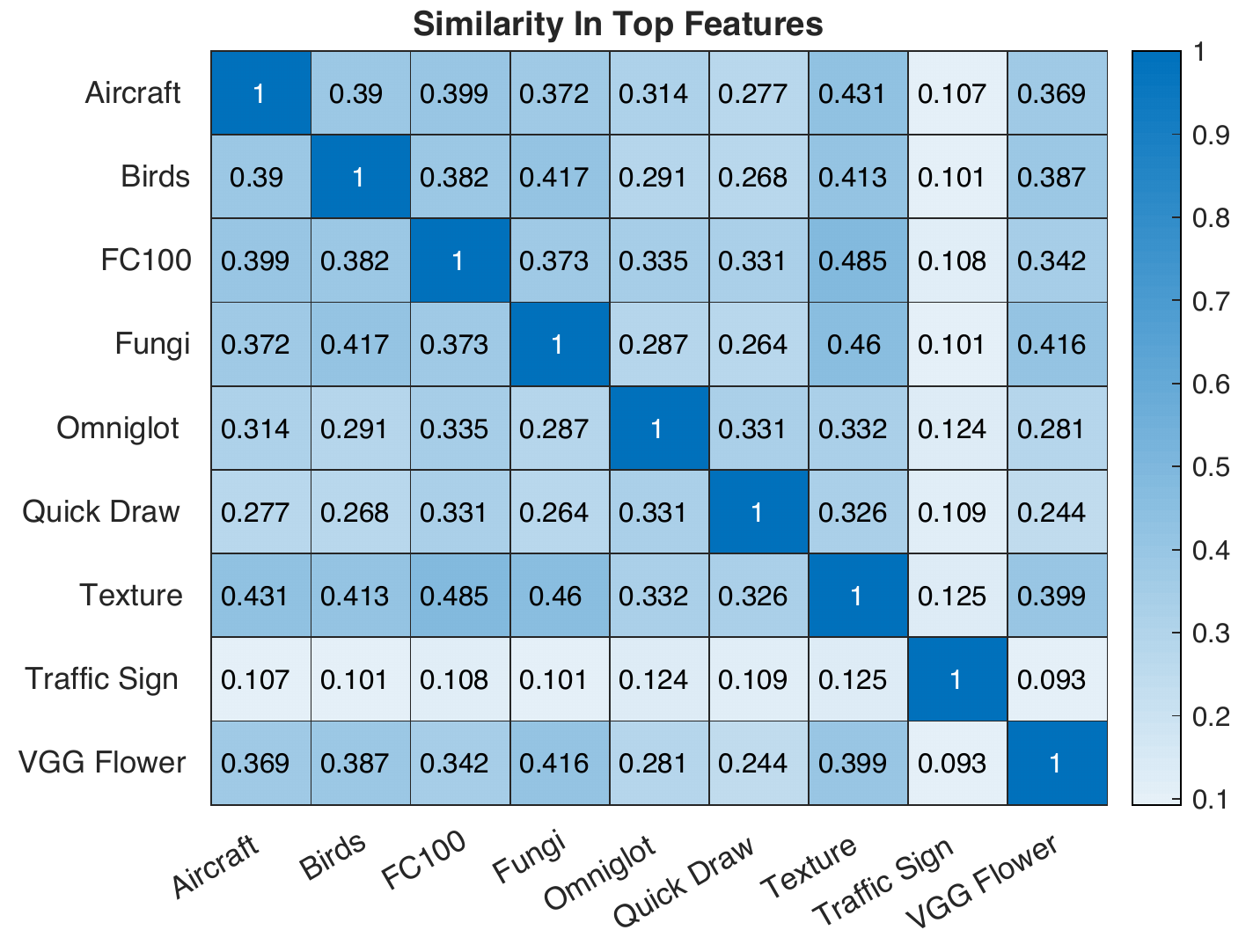} 
  \caption{Jaccard similarity of sets of most important features for the various (data set, data set) pairs.}
  \label{fig:top-features}
\end{figure}

\begin{figure}{}
  \centering
    \includegraphics[width=.99\linewidth]{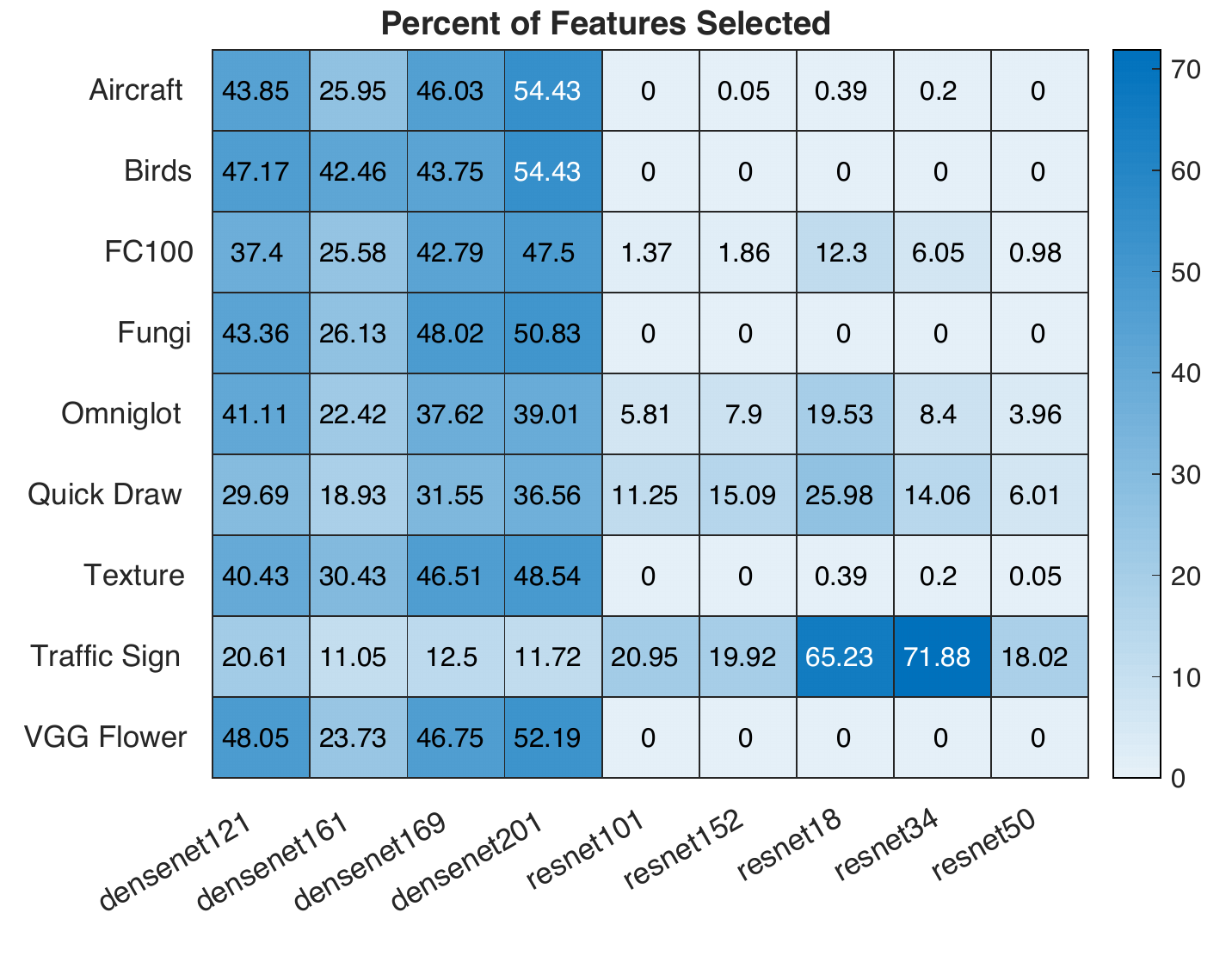} 
  \caption{Percent of each deep CNN's features that appear in the top 20\% of features, on each data set.}
  \label{fig:features-selected}
\end{figure}

For each of the (data set, data set) pairs, we compute the average Jaccard similarity of these top-feature sets.  Since each set consists of 20\% of the features, if each set of features chosen was completely random, for $n$ features in all, we would expect a Jaccard similarity of $\frac{0.04n}{.2n + .2n - 0.04n}$ $=$ $\frac{0.04}{.4 - 0.04}$ $=$ $0.111$. Anything greater indicates the sets of features selected are positively correlated; lower indicates a negative correlation.  Results are in Figure \ref{fig:top-features}.
For each of the nine CNNs, we also compute the fraction of each CNN's features that are in the top 20\% when a full library classifier is constructed. These percentages are in Figure \ref{fig:features-selected}.

The data in these two plots, along with the previous results, tells a consistent story: there appears to be little correspondence between data sets in terms of the set of features that are chosen as important across data sets.  The largest Jaccard value in Figure \ref{fig:top-features} is less than 0.5 (observed between FC100 and Texture).  This shows, in our opinion, a relatively weak correspondence.  The Traffic Sign data set, had an average Jaccard of $0.108$ across the other eight data sets, which is even lower than the $0.111$ that would be expected under a purely random feature selection regime. 
One might speculate that the lack of correspondence across data sets is evidence for the hypothesis that different tasks tend to use different features, which would explain why it is so effective to use an entire library of deep CNNs for few-shot learning.  



Also note that in Figure \ref{fig:features-selected}, we tend to see that different deep CNNs contribute ``important'' features at very different rates, depending on the particular few-shot classification task.  This also seems to be evidence that diversity is important.  In general, the DenseNets' features are preferred over the ResNets' features, but this is not universal, and there is a lot of variation. It may not be an accident that  the three data sets where the selection of features from library CNNs shows the most diversity in Figure \ref{fig:features-selected} (Traffic Sign, Quick Draw, and Omniglot) are the three data sets where the classifier built on top of all nine library CNNs has the largest advantage compared to the few-shot classifiers built on top of the single, ``Big Transfer''-based deep CNN.

\section{Related Work}

While most research on few-shot learning \cite{wang2018low, oreshkin2018tadam, rusu2018meta} has focused on developing new and sophisticated methods for learning a few-shot learner, there have recently been a few papers that, like this work, have suggested that transfer-based methods may be the most accurate.


In Section~\ref{sec:intro}, we mentioned Google's ``Big Transfer'' paper \cite{kolesnikov2019big}.  There, the authors argued that the best approach to few-shot learning is to concentrate on using a high-quality feature extractor rather than a sophisticated meta-learning approach.  We agree with this, but also give evidence that training a huge model and a massive data set may not be the only key to few-shot image classification. We found that a library with a wide diversity of high-quality deep CNNs can lead to substantially better accuracy than a single massive CNN, even when that deep CNN is trained on a much larger data set. Tian et al. \cite{tian2020rethinking} make a similar argument  to the Big Transfer authors, observing that a simple transfer-based approach can outperform a sophisticated meta-learner.

Chen et al. \cite{chen2019closer} were among the first researchers to point out that simple, transfer-based methods may outperform sophisticated few-shot learners.  They proposed Baseline which uses a simple linear classifier on top of a pre-trained deep CNN, and Baseline++ which uses a distance-based classifier.  Dhillon et al. \cite{dhillon2019baseline} also point out the utility of transfer-based methods, and propose a transductive fine-tuner; however, such a method relies on having an appropriately large number of relevant, unlabeled images to perform the transductive learning. Sun et al. \cite{sun2019meta} consider transfer learning, but where transfer is realized via shifting and scaling of the classification parameters.

Dvornik et al. \cite{2020arXiv200309338D} propose SUR, and argue that diversity in features is best obtained through diversity in data sets. Their method trains many feature extractors, one per data set. We argue that instead, a single high-quality data set is all that is needed. Adding additional feature extractors trained on that data set is the best way to higher accuracy. This may be counter-intuitive, but there is an obvious argument for this: training on less diverse data sets such as Aircraft or VGG Flowers is apt to result in feature extractors that are highly specialized, do not generalize well, and are not particularly useful. The proposed library-based classifier generally outperformed SUR in our experiments.

Dvornik et al. \cite{dvornik2019diversity} also consider the idea of ensembling for few-shot learning, although their idea is to simultaneously train a number of diverse deep CNNs as feature extractors during meta-learning; adding penalty terms that encourage both diversity and conformity during learning. We propose the simple idea of simply using a set of existing deep CNNs, trained by different groups of engineers.

\section{Conclusions}

We have examined the idea of using a library of deep CNNs as a basis for a high-quality few-shot learner.  We have shown that a learner built on top of a high-quality deep CNN can have remarkable accuracy, and that a learner built upon an entire library of CNNs can significantly outperform a few-shot learner built upon any one deep CNN. 

While we conjecture that it will be hard to do better than using a library of high-quality deep CNNs as the basis for a few-shot learner, there are key unanswered questions. First, future work should study if the accuracy continues to improve as the library is made even larger. Also, it may be expensive to feed a test image through a series of nine (or more) deep CNNs. It would be good to know if the computational cost of the library-based approach can be reduced.  Finally, it would be good to know if there are better methods to facilitate transfer than learning a simple MLP. 

\vspace{5 pt}
\noindent
\textbf{Acknowledgments.} The work in this paper was funded by NSF grant \#1918651; by NIH award \# UL1TR003167, and by ARO award \#78372-CS.

\balance

\clearpage
\newpage
\bibliography{main}
\bibliographystyle{icml2020}

\clearpage
\newpage
\section*{Appendix}
\beginsupplement
\section{Description of Data}
\subsection{Datasets}  
\begin{figure}[ht!]
    \centering
    \begin{subfigure}{.28\textwidth}
        \includegraphics[width=1\textwidth]{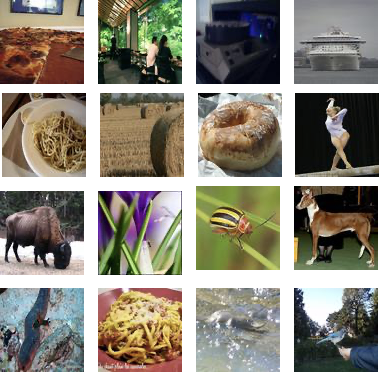}
        \caption{ILSVRC-2012}
        \label{fig:ilsvrc}
    \end{subfigure}
    \hskip10pt
    \begin{subfigure}{.28\textwidth}
        \includegraphics[width=1\textwidth]{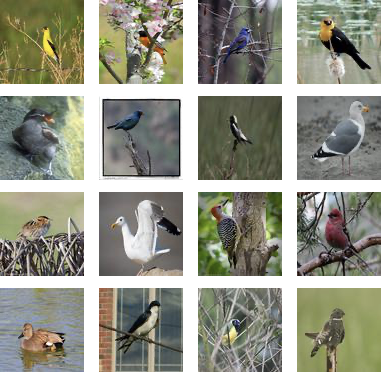}
        \caption{CUB-200-2011 Birds}
        \label{fig:cub}
    \end{subfigure}
    \hskip10pt
    \begin{subfigure}{.28\textwidth}
        \includegraphics[width=1\textwidth]{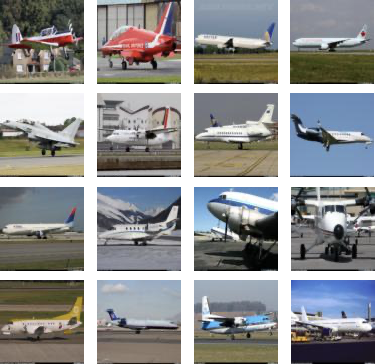}
        \caption{Aircraft}
        \label{fig:aircraft}
    \end{subfigure}
    \hskip10pt
    \begin{subfigure}{.28\textwidth}
        \includegraphics[width=1\textwidth]{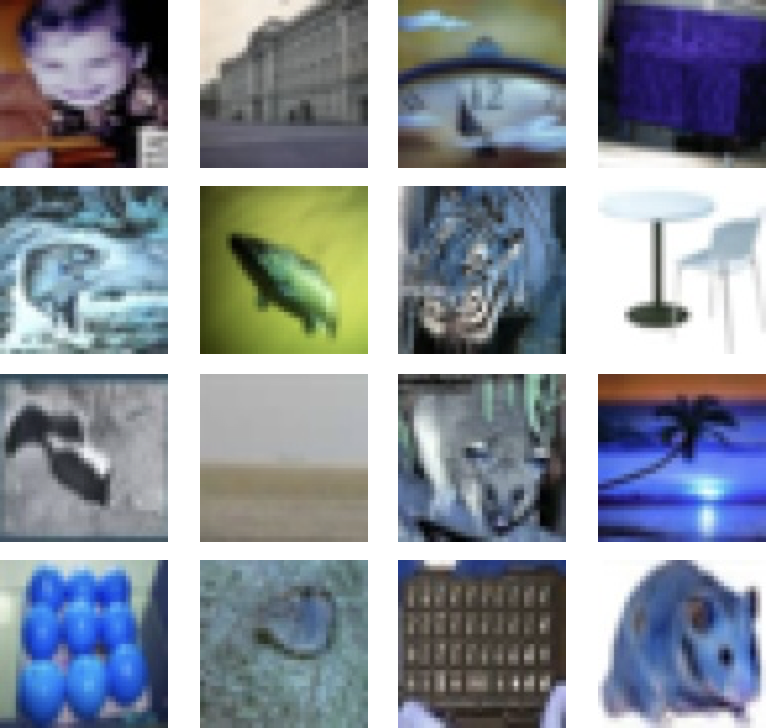}
        \caption{FC100}
        \label{fig:fc100}
    \end{subfigure}
    \hskip10pt
    \begin{subfigure}{.28\textwidth}
        \includegraphics[width=1\textwidth]{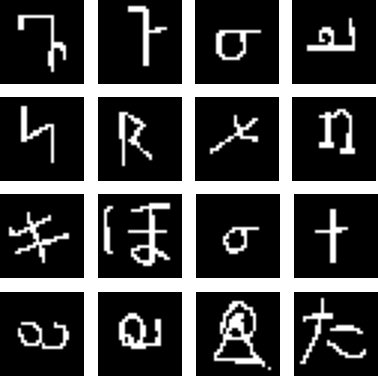}
        \caption{Omniglot}
        \label{fig:omniglot}
    \end{subfigure}
    \hskip10pt
    \begin{subfigure}{.28\textwidth}
        \includegraphics[width=1\textwidth]{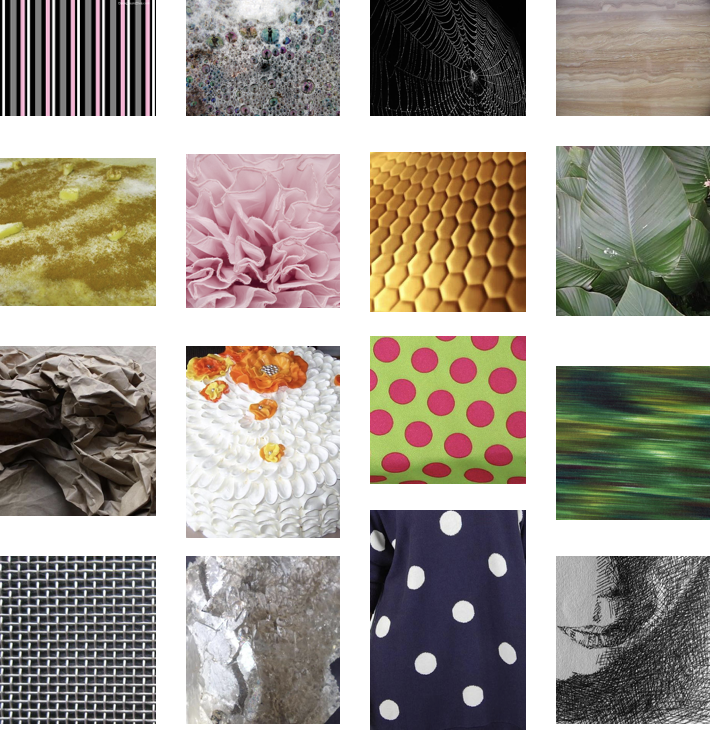}
        \caption{Texture}
        \label{fig:texture}
    \end{subfigure}
    \hskip10pt
    \begin{subfigure}{.28\textwidth}
        \includegraphics[width=1\textwidth]{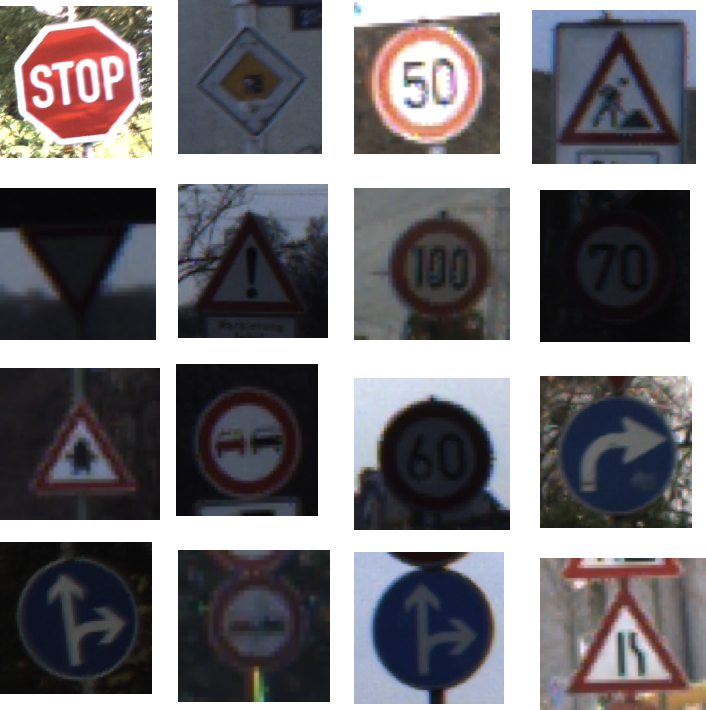}
        \caption{Traffic Sign}
        \label{fig:traffic}
    \end{subfigure}
    \hskip10pt
    \begin{subfigure}{.28\textwidth}
        \includegraphics[width=1\textwidth]{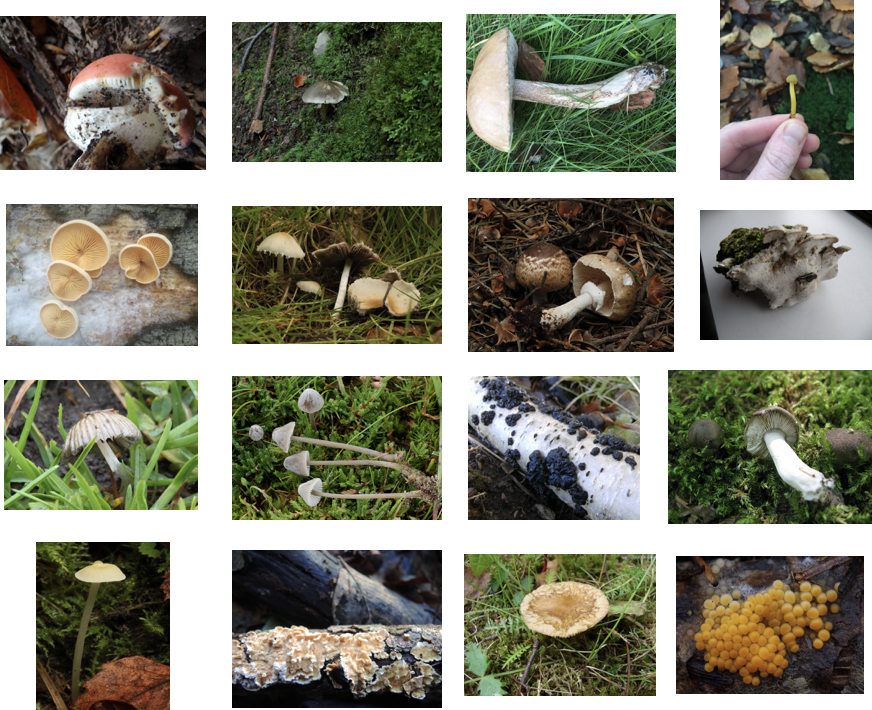}
        \caption{Fungi}
        \label{fig:fungi}
    \end{subfigure}
    \hskip10pt
    \begin{subfigure}{.28\textwidth}
        \includegraphics[width=1\textwidth]{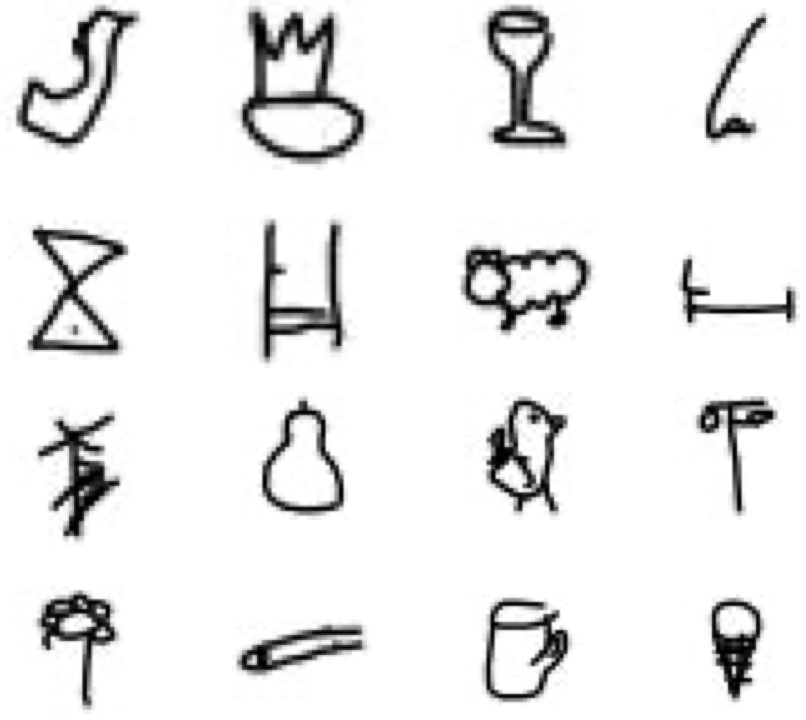}
        \caption{Quick Draw}
        \label{fig:quick_draw}
    \end{subfigure}
    \hskip10pt
    \begin{subfigure}{.28\textwidth}
        \includegraphics[width=1\textwidth]{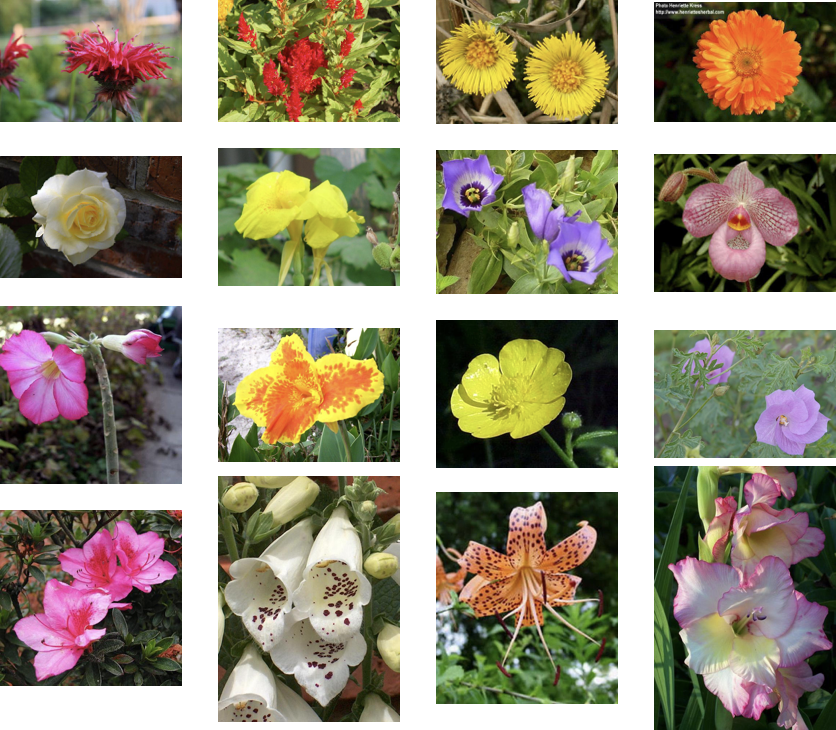}
        \caption{VGG Flower}
        \label{fig:vgg_flower}
    \end{subfigure}
    \caption{Images sampled from the data sets used in our experiments.}
\end{figure}

\noindent\textbf{ILSVRC2012 \cite{ILSVRC15}} Figure \ref{fig:ilsvrc}. A dataset of natural images of 1000 diverse categories, the most commonly used Imagenet dataset, primarily released for `Large Scale Visual Recognition Challenge'. We use the ILSVRC-2012 version as the original dataset for the classification challenge has not been modified since. The dataset has a little more than 1.2 million (1,281,167 to be precise) images with each class consisting of images ranging from 732 to 1300. \\

\noindent\textbf{CUB-200-2011 Birds \cite{WelinderEtal2010}} Figure \ref{fig:cub}. A dataset for fine-grained classification of 200 different bird species, an extended version of the CUB-200 dataset. The total number of images in the dataset is 11,788 with mostly 60 images per class.\\

\noindent\textbf{FGVC-Aircraft \cite{maji13fine-grained}} Figure \ref{fig:aircraft}. A dataset of images of aircrafts spanning 102 model variants with 10,200 total images and 100 images per class. \\

\noindent\textbf{FC100 \cite{oreshkin2018tadam}} Figure \ref{fig:fc100}. A dataset curated for few-shot learning based on the popular CIFAR100 \cite{krizhevsky2009learning} includes 100 classes and 600 32 × 32 color images per class. It offers a more challenging scenario with lower image resolution. \\

\noindent \textbf{Omniglot \cite{lake2015human}} Figure \ref{fig:omniglot}. A dataset of images of 1623 handwritten characters from 50 different alphabets. We consider each character as a separate class. The total number of images is 32,460 with 20 examples per class. \\

\noindent\textbf{Texture \cite{cimpoi14describing}} Figure \ref{fig:texture}. A dataset consists of 5640 images, organized according to 47 categories inspired from human perception. There are 120 images for each category. Image sizes range between 300x300 and 640x640, and the images contain at least 90\% of the surface representing the category attribute.\\

\noindent \textbf{Traffic Sign \cite{houben2013detection}} Figure \ref{fig:traffic}. A dataset of German Traffic Sign Recognition benchmark consisting of more than 50,000 images across 43 classes. \\

\noindent \textbf{FGCVx Fungi \cite{schroeder2018fgvcx}} Figure \ref{fig:fungi}. A dataset of wild mushrooms species which have been spotted and photographed by the general public in Denmark, containing over 100,000 images across 1,394 classes. \\

\noindent\textbf{Quick Draw \cite{jongejan2016quick}} Figure \ref{fig:quick_draw}. A dataset of 50 million drawings across 345 categories. We take the simplified drawings, which are 28x28 gray-scale bitmaps and aligned to the center of the drawing's bounding box. Considering the size of the full dataset, we randomly sample 1,000 images from each category.\\

\noindent\textbf{VGG  Flower \cite{nilsback2008automated}} Figure \ref{fig:vgg_flower}. A dataset consisting of 8189 images among 102 flower categories that commonly occuring in the United Kingdom. There are between 40 to 258 images in each category. \\

\subsection{Data Preparation} 
For all the datasets, we resize each image into 256 x 256 then crop 224 $\times$ 224 from the center (except quick draw which is already aligned to the center of the drawing's bounding box, so we directly resize quick draw images to 224 $\times$ 224). 

\subsection{Test Protocol}
In this work, for all the methods the training process is performed solely on ILSVRC dataset. For our library based methods, this is followed by hyperparameter validation on the CUB birds dataset. After that, each method is tested on the remaining eight datasets without further tuning.\\

To be more specific, for the library based methods we only use the pre-trained (on ILSVRC dataset) models. While for the meta-learning based methods, we randomly split ILSVRC into a base (training) set of 900 classes for meta-training and a validation set of the remaining 100 classes.\\

In order to evaluate a few-shot learner on a data set, for an ``$n$-way $m$-shot'' classification problem, we randomly select $n$ different classes, and then randomly select $m$ images from each class for training (equivalent to `support' images in meta-learning literature). We then randomly select $k$ images for rest of the images from each class for testing (equivalent to `query' images in meta-learning literature). We perform this evaluation for $n$ in $\{5, 20, 40\}$ and $m$ in $\{1, 5\}$.  \\

For the library based methods, the query size $k$ is set to 15 (except FGCVx Fungi dataset). For the meta-learning based methods, due to GPU memory constraints, for each class in a task we used 15 query images for 5-way, 10 query

\onecolumn
\twocolumn

\noindent images for 10-way, and 5 query images for 40-way problems, with only exception being the Fungi dataset. Fungi dataset has several classes with a very small number of images (6 being the minimum). Hence for Fungi dataset, we use 5 query images per class for 1-shot and 1 query image per class for 5-shot problems. Finally, for every problem, we report the mean of 600 randomly generated test tasks along with the 95\% confidence intervals.

For the FC100 data set, there is a small portion overlaps  with ILSVRC2012 data set but we still think that the FC100 as a testing set is interesting, as (1) all of the few-shot learners benefitted from the overlapping classes,  and (2) this shows how the methods work in the case that the test classes are close to something that the few-shot learner has seen before.

\section{Hyper-parameters for Library-Based \\Learners}
In order to achieve an ``out-of-the-box'' few-shot learner that can be used on any (very small) training set $D_{few}$ without additional data or knowledge of the underlying distribution, an extensive hyper-parameter search is performed on the validation dataset (CUB birds). The hyper-parameters that are found to be the best are then applied in the few-shot training and testing phase for each of our library based methods.  The hyper-parameters we are considering include learning rate, number of training epochs, L2-regularization penalty constant, the number of neurons in the MLP hidden layer, and whether to drop the hidden layer altogether.  A separate hyper-parameter search is used for 5-way, 20-way, and 40-way 1-shot classification. 5-shot problems are using the same hyper-parameters as 1-shot ones. Experiments suggest that dropping the hidden layer did not significantly help performance, but did occasionally hurt performance significantly on the validation set; as a result, we always use a hidden layer. We train all our methods using Adam optimizer. The hyper-parameters details can be found in Table \ref{tab:hyper_parameter}.

\section{Competitive Methods}
For methods that require a pre-trained CNN (FEAT, Meta-transfer, and SUR), we use the pre-trained ResNet18 pytorch library as the backbone. We follow the hyper-parameter setting from \cite{sun2019meta}, \cite{ye2020few}, and \cite{2020arXiv200309338D}. For the FEAT and Meta-transfer methods, we perform meta-training on ILSVRC dataset\cite{ILSVRC15} before testing on eight different datasets. For the SUR method, we follow \cite{2020arXiv200309338D} and build a multi-domain representation by pre-training multiple ResNet18 on Meta-Dataset \cite{2019arXiv190303096T} (one  per  data  set). To evaluate SUR on data set $X$, we use feature extractors trained on the rest of the data sets in \{Omniglot, Aircraft, Birds, Texture, Quickdraw, Flowers, Fungi,and ILSVRC\}. Traffic Sign and FC100 data sets are reserved for testing only. To be more specific, the meta-training setting are as follows: \\

\noindent \textbf{Meta-transfer Learning}\\The base-learner is optimized by batch gradient descent with the learning rate of $10^{-2}$ and gets updated every 50 steps. The meta learner is optimized by Adam optimizer with an initial learning rate of $10^{-3}$, and decaying 50\% every 1,000 iterations until $10^{-4}$.  \\

\noindent \textbf{FEAT}\\The vanilla stochastic gradient descent with Nesterov acceleration is applied. The initial rate is set to $2 \times10^{-4}$ and decreases every 40 steps with a decay rate of $5 \times10^{-4}$ and momentum of 0.9. The learning rate for the scheduler is set to 0.5. \\

\noindent \textbf{SUR}\\We follow \cite{2020arXiv200309338D} and apply SGD with momentum during optimization. The learning rate is adjusted with cosine annealing. The initial learning rate, the maximum number of training iterations (“Max Iter.”) and annealing frequency (“Annealing Freq.”) are adjusted individually according to each data set (Table \ref{tab:SUR_hyper_parameter}). Data augmentation is also deployed to regularize the training process, which includes random crops and random color augmentations with a constant weight decay of $7 \times10^{-4}$.\\
\begin{table*}[t!]
\centering
\begin{small}

\begin{tabular} {l|ccccr}

\toprule
     \multicolumn{1}{c}{} & {Learning} & Weight & Max & Annealing &\multicolumn{1}{c}{Batch}\\
      \multicolumn{1}{c}{} & Rate & Decay & Iter. & Freq. & Size \\
\midrule
\midrule
     ILSVRC   & $3 \times 10^{-2}$ & $7 \times 10^{-4}$ & 480,000 & 48,000 & 64 \\
     Omniglot & $3 \times 10^{-2}$ & $7 \times 10^{-4}$ & 50,000 & 3,000 & 16 \\
     Aircraft & $3 \times 10^{-2}$ & $7 \times 10^{-4}$ & 50,000 & 3,000 & 8 \\
     Birds & $3 \times 10^{-2}$ & $7 \times 10^{-4}$ & 50,000 & 3,000 & 16 \\
     Textures & $3 \times 10^{-2}$ & $7 \times 10^{-4}$ & 50,000 & 1,500 & 32 \\
     Quick Draw & $1 \times 10^{-2}$ & $7 \times 10^{-4}$ & 480,000 & 48,000 & 64 \\
     Fungi & $3 \times 10^{-2}$ & $7 \times 10^{-4}$ & 480,000 & 15,000 & 32 \\
     VGG Flower & $3 \times 10^{-2}$ & $7 \times 10^{-4}$ & 50,000 & 1,500 & 8 \\
\midrule
\bottomrule
\end{tabular}
\end{small}
\vspace{5 pt}
\caption{Hyper-parameter settings for SUR individual feature networks on MetaDataset.}
\label{tab:SUR_hyper_parameter}
\end{table*}

\noindent \textbf{Baselines}\\
We train the two pre-training based methods, Baseline and Baseline++ \cite{chen2019closer} following the hyper-parameters suggested by the original authors. However, since we train them on ILSVRC data as opposed to mini-imagenet \cite{ravi2016optimization}. During the training stage, we train 50 epochs with a batch size of 16. In the paper, the authors have trained 400 epochs on the base set of mini-imagenet consisting of 64 classes. Mini-imagenet has 600 images per class, whereas ILSVRC has an average of around 1,200 images per class. So, the total number of batches trained in our baselines is $50\times(1000\times1,200)/16 = 3,750,000$, as opposed to $400\times(64\times600)/16 = 960,000$ in the original paper.\\

\begin{table*}[ht!]
\centering
\begin{small}
\begin{tabular}{l|cccccccr} 

\toprule
     \multicolumn{1}{c}{} & Aircraft & FC100 & Omniglot & Texture & Traffic & Fungi & Quick Draw & \multicolumn{1}{c}{VGG Flower} \\ 

\midrule
     Baseline & 36.6 $\pm$ 0.7 & 40.3 $\pm$ 0.8 & 65.2 $\pm$ 1.0 & 42.9 $\pm$ 0.7 & 57.6 $\pm$ 0.9 & 35.9 $\pm$ 0.9 & 49.0 $\pm$ 0.9 & 67.9 $\pm$ 0.9 \\

     Baseline++ & 33.9 $\pm$ 0.7 & 39.3 $\pm$ 0.7 & 59.8 $\pm$ 0.7 & 40.8 $\pm$ 0.7 & 58.8 $\pm$ 0.8 & 35.4 $\pm$ 0.9 & 46.6 $\pm$ 0.8 & 58.6 $\pm$ 0.9\\

     MAML & 26.5 $\pm$ 0.6 & 39.4 $\pm$ 0.8 & 50.7 $\pm$ 1.0 & 38.1 $\pm$ 0.9 & 45.6 $\pm$ 0.8 & 34.8 $\pm$ 1.0 & 46.0 $\pm$ 1.0 & 54.7 $\pm$ 0.9\\

     MatchingNet & 29.9 $\pm$ 0.6 & 38.2 $\pm$ 0.8 & 51.7 $\pm$ 1.0 & 39.3 $\pm$ 0.7 & 55.2 $\pm$ 0.8 & 38.1 $\pm$ 0.9 & 50.2 $\pm$ 0.8 & 51.8 $\pm$ 0.9\\

     ProtoNet & 31.8 $\pm$ 0.6 & 40.9 $\pm$ 0.8 & 79.2 $\pm$ 0.8 & 42.0 $\pm$ 0.8 & 54.4 $\pm$ 0.9 & 36.7 $\pm$ 0.9 & 55.8 $\pm$ 1.0 & 59.1 $\pm$ 0.9\\

     RelationNet & 31.2 $\pm$ 0.6 & 46.3 $\pm$ 0.9 & 69.6 $\pm$ 0.9 & 41.0 $\pm$ 0.8 & 54.6 $\pm$ 0.8 & 36.8 $\pm$ 1.0 & 52.5 $\pm$ 0.9 & 55.5 $\pm$ 0.9\\

     Meta-transfer & 30.4 $\pm$ 0.6  & 57.6 $\pm$ 0.9 & 78.9 $\pm$ 0.8 & 50.1 $\pm$ 0.8 & 62.3 $\pm$ 0.8 & 45.8 $\pm$ 1.0 & 58.4 $\pm$ 0.9 & 73.2 $\pm$ 0.9 \\

     FEAT & 33.2 $\pm$ 0.7  & 42.1 $\pm$ 0.8 & 69.8 $\pm$ 0.9 & 51.8 $\pm$ 0.9 & 49.0 $\pm$ 0.8 & 46.9 $\pm$ 1.0 & 53.1 $\pm$ 0.8 & 75.3 $\pm$ 0.9 \\
     
     SUR & 33.5 $\pm$ 0.6  & 42.1 $\pm$ 1.0 & 93.4 $\pm$ 0.5 & 42.8 $\pm$ 0.8 & 45.3 $\pm$ 0.9 & 44.1 $\pm$ 1.0 & 54.3 $\pm$ 0.9 & 72.6 $\pm$ 1.0 \\
\midrule
     Worst library- & 40.9 $\pm$ 0.9 & 50.8 $\pm$ 0.9& 77.2 $\pm$ 0.8 & 59.0 $\pm$ 0.9 & 55.5 $\pm$ 0.8 & 53.0 $\pm$ 0.9 & 57.3 $\pm$ 0.9 & 79.7 $\pm$ 0.8 \\
     based & RN18 & DN121 & RN152 & DN169 & RN152 & DN201 & RN101 & RN18 \\

     Best library- & \textbf{46.1 $\pm$ 1.0} & \textbf{61.2 $\pm$ 0.9} & \textbf{86.5 $\pm$ 0.6} & \textbf{65.1 $\pm$ 0.9} & \textbf{66.6 $\pm$ 0.9} & \textbf{56.6 $\pm$ 0.9} & \textbf{62.8 $\pm$ 0.9} & \textbf{83.2 $\pm$ 0.8} \\
     based & DN161 & RN152 & DN121 & RN101 & DN201 & DN121 & RN18 & DN201 \\
\bottomrule

\end{tabular}
\end{small}
\vspace{5 pt}
\caption{Comparing competitive methods with the simple library-based learners, on the 5-way, 1-shot problem.}
\label{tab:comparison-5-1}
\end{table*}

\noindent \textbf{Metric-Learning Methods and MAML}\\
For the three most popular metric-learning methods, MatchingNet \cite{vinyals2016matching}, ProtoNet \cite{snell2017prototypical} and RelationNet \cite{sung2018learning}, again we followed the implementation and hyper-parameters provided by \cite{chen2019closer}. 

All the metric-learning methods and MAML \cite{finn2017model} are trained using the Adam optimizer with initial learning rate of $10^{-3}$. For MAML, the inner learning rate is kept at $10^{-2}$ as suggested by the original authors. And following \cite{chen2019closer}, we do the following modifications: For MatchingNet, we use an FCE classification layer without fine-tuning and also multiply the cosine similarity by a constant scalar. For RelationNet, we replace the L2 norm with a softmax layer to expedite training. For MAML, we use a first-order approximation in the gradient for memory efficiency. Even then, we could not train MAML for 40-way in a single GPU due to memory shortage. Hence, we drop MAML for 40-way experiments.

During the meta-training stage of all these methods, we train 150,000 episodes for 5-way 1-shot, 5-way 5-shot, and 20-way 1-shot problems, and 50,000 episodes for all other problems on the base split of ILSVRC. Here an episode refers to one meta-learning task that includes training on the `support' images and testing on the `query' images. The stopping episode number is chosen based on no significant increase in validation accuracy. In the original paper, the authors trained 60,000 episodes for 1-shot and 40,000 episodes for 5-shot tasks. We meticulously observe that those numbers are too low for certain problems in terms of validation accuracy. Hence, we allow more episodes. We select the best accuracy model on the validation set of ILSVRC for meta-testing on other datasets.

\section{Complete Results}
Here we conduct additional experiments which are not reported in our main paper.\\

\begin{table*}[ht!]
\centering
\begin{small}
\begin{tabular}{l|cccccccr} 

\toprule
     \multicolumn{1}{c}{} & Aircraft & FC100 & Omniglot & Texture & Traffic & Fungi & Quick Draw & \multicolumn{1}{c}{VGG Flower} \\ 

\midrule

     Baseline & 11.4 $\pm$ 0.2 & 15.4 $\pm$ 0.3 & 38.9 $\pm$ 0.6 & 17.6 $\pm$ 0.3 & 27.8 $\pm$ 0.4 & 13.1 $\pm$ 0.3 & 21.9 $\pm$ 0.4 & 38.8 $\pm$ 0.4 \\

     Baseline++ & 10.1 $\pm$ 0.2 & 15.8 $\pm$ 0.3 & 39.2 $\pm$ 0.4 & 18.1 $\pm$ 0.3 & 31.5 $\pm$ 0.3 & 13.7 $\pm$ 0.3 & 22.5 $\pm$ 0.3 & 33.1 $\pm$ 0.4\\

     MAML & 7.6 $\pm$ 0.1 & 17.2 $\pm$ 0.3 & 29.1 $\pm$ 0.4 & 14.8 $\pm$ 0.2 & 19.4 $\pm$ 0.3 & 11.5 $\pm$ 0.3 & 19.7 $\pm$ 0.3 & 21.8 $\pm$ 0.3\\

     MatchingNet & 7.7 $\pm$ 0.1 & 17.0 $\pm$ 0.3 & 44.6 $\pm$ 0.5 & 19.8 $\pm$ 0.3 & 26.2 $\pm$ 0.3 & 16.1 $\pm$ 0.4 & 26.7 $\pm$ 0.4 & 29.9 $\pm$ 0.4\\

     ProtoNet & 11.5 $\pm$ 0.2 & 20.1 $\pm$ 0.3 & 58.8 $\pm$ 0.5 & 20.0 $\pm$ 0.3 & 29.5 $\pm$ 0.4 & 16.2 $\pm$ 0.4 & 30.7 $\pm$ 0.4 & 40.7 $\pm$ 0.4\\

     RelationNet & 11.0 $\pm$ 0.2 & 18.7 $\pm$ 0.3 & 51.3 $\pm$ 0.5 & 18.6 $\pm$ 0.3 & 28.5 $\pm$ 0.3 & 15.9 $\pm$ 0.4 & 29.1 $\pm$ 0.4 & 35.5 $\pm$ 0.4\\

     Meta-transfer & 12.8 $\pm$ 0.2  & 30.9 $\pm$ 0.4 & 58.6 $\pm$ 0.5 & 27.2 $\pm$ 0.4 & 35.8 $\pm$ 0.4 & 23.1 $\pm$ 0.4 & 35.2 $\pm$ 0.4 & 52.0 $\pm$ 0.5 \\
     
     FEAT & 15.1 $\pm$ 0.3  & 18.9 $\pm$ 0.4 & 51.1 $\pm$ 0.6 & 28.9 $\pm$ 0.5 & 31.7 $\pm$ 0.4 & 25.7 $\pm$ 0.4 & 28.7 $\pm$ 0.5 & 56.5 $\pm$ 0.6 \\
     
     SUR & 14.2 $\pm$ 0.3  & 19.1 $\pm$ 0.4 & 84.2 $\pm$ 0.4 & 21.0 $\pm$ 0.3 & 26.2 $\pm$ 0.3 & 21.7 $\pm$ 0.4 & 34.2 $\pm$ 0.4 & 57.1 $\pm$ 0.5 \\
\midrule
     Worst library- & 20.1 $\pm$ 0.3 & 27.8 $\pm$ 0.4 & 56.2 $\pm$ 0.5 & 38.0 $\pm$ 0.4 & 29.7 $\pm$ 0.3 & 31.7 $\pm$ 0.4 & 33.3 $\pm$ 0.5 & 62.4 $\pm$ 0.5 \\
     based & RN101 & DN121 & RN101 & RN18 & RN101 & RN101 & RN101 & RN101 \\

     Best library- & \textbf{24.3 $\pm$ 0.3} & \textbf{36.4 $\pm$ 0.4} & \textbf{69.1 $\pm$ 0.4} & \textbf{42.5 $\pm$ 0.4} & \textbf{38.5 $\pm$ 0.4} & \textbf{33.9 $\pm$ 0.5} & \textbf{39.5 $\pm$ 0.5} & \textbf{70.0 $\pm$ 0.5} \\
     based & DN161 & RN152 & DN121 & DN152 & DN201 & DN161 & DN201 & DN161 \\
\bottomrule

\end{tabular}
\end{small}
\vspace{5 pt}
\caption{Comparing competitive methods with the simple library-based learners, on the 20-way, 1-shot problem.}
\label{tab:comparison-20-1}
\end{table*}

\begin{table*}[ht!]
\centering
\begin{small}
\begin{tabular}{l|cccccccr} 

\toprule
     \multicolumn{1}{c}{} & Aircraft & FC100 & Omniglot & Texture & Traffic & Fungi & Quick Draw & \multicolumn{1}{c}{VGG Flower} \\ 

\midrule

     Baseline & 6.9 $\pm$ 0.2 & 9.7 $\pm$ 0.1 & 27.1 $\pm$ 0.4 & 11.2 $\pm$ 0.1 & 20.0 $\pm$ 0.3 & 8.2 $\pm$ 0.2 & 14.8 $\pm$ 0.3 & 28.7 $\pm$ 0.3 \\

     Baseline++ & 6.1 $\pm$ 0.1 & 10.1 $\pm$ 0.1 & 29.9 $\pm$ 0.3 & 12.1 $\pm$ 0.2 & 23.2 $\pm$ 0.2 & 8.7 $\pm$ 0.2 & 15.9 $\pm$ 0.2 & 24.2 $\pm$ 0.3\\

     MatchingNet & 5.1 $\pm$ 0.1 & 11.6 $\pm$ 0.2 & 32.5 $\pm$ 0.4 & 15.0 $\pm$ 0.2 & 19.3 $\pm$ 0.2 & 11.0 $\pm$ 0.2 & 19.0 $\pm$ 0.3 & 26.9 $\pm$ 0.3\\

     ProtoNet & 7.2 $\pm$ 0.2 & 13.8 $\pm$ 0.2 & 47.2 $\pm$ 0.4 & 14.4 $\pm$ 0.2 & 21.6 $\pm$ 0.3 & 11.3 $\pm$ 0.2 & 22.9 $\pm$ 0.3 & 31.0 $\pm$ 0.3\\

     RelationNet & 6.2 $\pm$ 0.2 & 13.8 $\pm$ 0.2 & 45.2 $\pm$ 0.4 & 11.4 $\pm$ 0.2 & 18.2 $\pm$ 0.2 & 10.7 $\pm$ 0.2 & 21.1 $\pm$ 0.3 & 28.1 $\pm$ 0.3\\

     Meta-transfer & 7.6 $\pm$ 0.2  & 20.6 $\pm$ 0.2 & 37.5 $\pm$ 0.3 & 19.2 $\pm$ 0.2 & 24.4 $\pm$ 0.2 & 10.2 $\pm$ 0.2 & 22.2 $\pm$ 0.2 & 40.4 $\pm$ 0.3 \\
     
     FEAT & 10.1 $\pm$ 0.2  & 12.9 $\pm$ 0.3 & 35.7 $\pm$ 0.4 & 21.8 $\pm$ 0.3 & 24.9 $\pm$ 0.3 & 18.0 $\pm$ 0.3 & 19.4 $\pm$ 0.3 & 47.6 $\pm$ 0.4 \\

     SUR & 9.9 $\pm$ 0.1  & 13.3 $\pm$ 0.2 & 78.1 $\pm$ 0.3 & 15.0 $\pm$ 0.2 & 19.7 $\pm$ 0.2 & 15.6 $\pm$ 0.3 & 26.7 $\pm$ 0.3 & 48.8 $\pm$ 0.3 \\
\midrule
     Worst library- & 14.2 $\pm$ 0.2 & 19.6 $\pm$ 0.2 & 47.3 $\pm$ 0.3 & 28.8 $\pm$ 0.2 & 22.2 $\pm$ 0.2 & 23.7 $\pm$ 0.3 & 26.4 $\pm$ 0.3 & 53.1 $\pm$ 0.3 \\
     based & RN34 & RN18 & RN152 & RN18 & RN152 & RN34 & RN152 & RN34 \\

     Best library- & \textbf{17.4 $\pm$ 0.2} & \textbf{27.2 $\pm$ 0.3} & \textbf{61.6 $\pm$ 0.3} & \textbf{33.2 $\pm$ 0.3} & \textbf{29.5 $\pm$ 0.2} & \textbf{26.8 $\pm$ 0.3} & \textbf{31.2 $\pm$ 0.3} & \textbf{62.8 $\pm$ 0.3} \\
     based & DN161 & RN152 & DN201 & DN152 & DN201 & DN161 & DN201 & DN161 \\
\bottomrule

\end{tabular}
\end{small}
\vspace{5 pt}
\caption{Comparing competitive methods with the simple library-based learners, on the 40-way, 1-shot problem.}
\label{tab:comparison-40-1}
\end{table*}

\begin{table*}[t!]
\centering
\begin{small}
\begin{tabular}{l|cccccccr} 

\toprule
     \multicolumn{1}{c}{} & Aircraft & FC100 & Omniglot & Texture & Traffic & Fungi & Quick Draw & \multicolumn{1}{c}{VGG Flower} \\ 

\midrule
     Baseline & 17.0 $\pm$ 0.2 & 29.4 $\pm$ 0.3 & 80.0 $\pm$ 0.3 & 27.1 $\pm$ 0.2 & 49.1 $\pm$ 0.3 & 24.0 $\pm$ 0.5 & 41.5 $\pm$ 0.3 & 68.3 $\pm$ 0.3 \\

     Baseline++ & 12.3 $\pm$ 0.2 & 24.4 $\pm$ 0.3 & 67.0 $\pm$ 0.3 & 25.1 $\pm$ 0.3 & 44.1 $\pm$ 0.3 & 19.7 $\pm$ 0.5 & 35.2 $\pm$ 0.3 & 53.9 $\pm$ 0.3\\

     MatchingNet & 8.6 $\pm$ 0.2 & 22.1 $\pm$ 0.3 & 59.1 $\pm$ 0.4 & 23.3 $\pm$ 0.3 & 37.1 $\pm$ 0.3 & 19.0 $\pm$ 0.5 & 31.5 $\pm$ 0.3 & 46.7 $\pm$ 0.4\\

     ProtoNet & 13.8 $\pm$ 0.2 & 28.0 $\pm$ 0.3 & 80.0 $\pm$ 0.3 & 29.6 $\pm$ 0.3 & 39.7 $\pm$ 0.3 & 23.6 $\pm$ 0.5 & 42.6 $\pm$ 0.4 & 61.6 $\pm$ 0.3\\

     RelationNet & 10.7 $\pm$ 0.2 & 26.1 $\pm$ 0.3 & 77.0 $\pm$ 0.3 & 18.7 $\pm$ 0.2 & 29.6 $\pm$ 0.3 & 18.3 $\pm$ 0.5 & 40.6 $\pm$ 0.3 & 47.0 $\pm$ 0.3\\

     Meta-transfer & 9.7 $\pm$ 0.2  & 29.9 $\pm$ 0.2 & 48.1 $\pm$ 0.3 & 29.8 $\pm$ 0.2 & 33.3 $\pm$ 0.2 & 12.2 $\pm$ 0.3 & 31.6 $\pm$ 0.2 & 55.4 $\pm$ 0.3 \\
     
     FEAT & 16.2 $\pm$ 0.3  & 27.1 $\pm$ 0.4 & 58.5 $\pm$ 0.4 & 36.8 $\pm$ 0.3 & 37.3 $\pm$ 0.3 & 32.9 $\pm$ 0.6 & 35.6 $\pm$ 0.4 & 74.0 $\pm$ 0.4 \\

     SUR & 15.5 $\pm$ 0.2  & 32.6 $\pm$ 0.3 & 94.2 $\pm$ 0.1 & 25.8 $\pm$ 0.1 & 37.0 $\pm$ 0.2 & 26.3 $\pm$ 0.6 & 45.0 $\pm$ 0.3 & 69.8 $\pm$ 0.3 \\
\midrule
     Worst library- & 28.4 $\pm$ 0.2 & 37.3 $\pm$ 0.2 & 79.9 $\pm$ 0.3 & 48.4 $\pm$ 0.2 & 47.2 $\pm$ 0.2 & 46.6 $\pm$ 0.3 & 49.8 $\pm$ 0.3 & 81.4 $\pm$ 0.2 \\
     based & RN34 & RN18 & RN152 & RN18 & RN152 & RN34 & RN152 & RN34 \\

     Best library- & \textbf{35.9 $\pm$ 0.2} & \textbf{48.2 $\pm$ 0.3} & \textbf{89.4 $\pm$ 0.2} & \textbf{55.4 $\pm$ 0.2} & \textbf{57.5 $\pm$ 0.2} & \textbf{52.1 $\pm$ 0.3} & \textbf{55.5 $\pm$ 0.3} & \textbf{88.9 $\pm$ 0.2} \\
     based & DN161 & RN152 & DN201 & DN161 & DN201 & DN161 & DN201 & DN161 \\
\bottomrule

\end{tabular}
\end{small}
\vspace{5 pt}
\caption{Comparing competitive methods with the simple library-based learners, on the 40-way, 5-shot problem.}
\label{tab:comparison-40-5}
\end{table*}

\noindent \textbf{Single Library Learners VS. Competitive Methods}\\
Table \ref{tab:comparison-5-1}, \ref{tab:comparison-20-1}, \ref{tab:comparison-40-1}, \ref{tab:comparison-40-5} show the performance comparison of single library learners and the competitive methods including Baseline, Baseline++, MAML, MatchingNet, ProtoNet, RelationNet, Meta-transfer Learning, and FEAT. The comparison is conducted on problems of 1-shot under 5-way, 20-way, and 40-way and 5-shot under 40-way.\\

\noindent \textbf{Full Library VS. Google BiT Methods}\\
Table \ref{tab:vs_bigt_1shot} shows the performance comparison of the full library method and the Google BiT methods. The problems addressed here are 1-shot under 5-way, 20-way, and 40-way. The full library method utilizes a library of nine, ILSVRC-trained CNNs, while the Google BiT methods individually use three deep CNNs trained on the full ImageNet. \\

\noindent \textbf{Full Library VS. Hard and Soft Ensemble Methods}\\
Table \ref{tab:ensemble_1shot} compares the performances of the full library method and the hard and soft ensemble methods (bagging) for 1-shot problems under 5-way, 20-way, and 40-way.

\begin{table*}[t!]
\centering
\begin{small}

\begin{tabular} {l|cccccccr}

\toprule
     \multicolumn{1}{c}{} & Aircraft & FC100 & Omniglot & Texture & Traffic & Fungi & QDraw & \multicolumn{1}{c}{Flower} \\
\midrule
     \multicolumn{1}{c}{} & \multicolumn{8}{c}{5-way, 1-shot} \\
\midrule
     Full Library & 44.9 $\pm$ 0.9 & 60.9 $\pm$ 0.9 & \textbf{88.4 $\pm$ 0.7} & 68.4 $\pm$ 0.8 & \textbf{66.2 $\pm$ 0.9} & 57.7 $\pm$ 1.0 & \textbf{65.4 $\pm$ 0.9} & 86.3 $\pm$ 0.8\\
     BiT-ResNet-101-3& 42.2 $\pm$ 1.0 & 58.5 $\pm$ 0.9 & 76.2 $\pm$ 1.1 & 63.6 $\pm$ 0.9 & 47.4 $\pm$ 0.8 & \textbf{63.7 $\pm$ 0.9} & 54.9 $\pm$ 0.9 & 98.0 $\pm$ 0.3 \\
     BiT-ResNet-152-4& 40.5 $\pm$ 0.9 & 61.0 $\pm$ 0.9 & 78.4 $\pm$ 0.9 & 65.6 $\pm$ 0.8 & 48.4 $\pm$ 0.8 & 62.2 $\pm$ 1.0 & 55.4 $\pm$ 0.9 &  97.9 $\pm$ 0.3 \\
     BiT-ResNet-50-1& \textbf{45.0 $\pm$ 1.0} & \textbf{61.9 $\pm$ 0.9} & 79.4 $\pm$ 0.9 & \textbf{68.5 $\pm$ 0.9} & 56.1 $\pm$ 0.8  & 61.6 $\pm$ 1.0 & 54.0 $\pm$ 0.9 & \textbf{98.5 $\pm$ 0.2} \\
\midrule
     \multicolumn{1}{c}{} & \multicolumn{8}{c}{20-way, 1-shot} \\
\midrule 
     Full Library & \textbf{25.2 $\pm$ 0.3} & 36.8 $\pm$ 0.4 & \textbf{76.4 $\pm$ 0.4} & \textbf{46.3 $\pm$ 0.4} & \textbf{40.0 $\pm$ 0.4} & 37.6 $\pm$ 0.5 & \textbf{44.2 $\pm$ 0.5} & 75.5 $\pm$ 0.5\\
     BiT-ResNet-101-3& 19.9 $\pm$ 0.3 & 34.5 $\pm$ 0.4 & 53.4 $\pm$ 0.6 & 44.2 $\pm$ 0.4 & 24.6 $\pm$ 0.3 & \textbf{41.6 $\pm$ 0.5} & 32.0 $\pm$ 0.4 & \textbf{95.7 $\pm$ 0.2} \\
     BiT-ResNet-152-4& 18.2 $\pm$ 0.3 & \textbf{37.2 $\pm$ 0.4} & 51.0 $\pm$ 0.6 & 44.3 $\pm$ 0.4 & 23.6 $\pm$ 0.3 & 40.3 $\pm$ 0.5 & 29.0 $\pm$ 0.4 &  95.0 $\pm$ 0.2 \\
     BiT-ResNet-50-1& 22.6 $\pm$ 0.4 & 36.1 $\pm$ 0.4 & 58.1 $\pm$ 0.5 & 45.7 $\pm$ 0.4 & 28.3 $\pm$ 0.3 & 39.2 $\pm$ 0.4 & 30.0 $\pm$ 0.4 &   95.4 $\pm$ 0.2 \\
\midrule
     \multicolumn{1}{c}{} & \multicolumn{8}{c}{40-way, 1-shot} \\
\midrule 
     Full Library & \textbf{18.6 $\pm$ 0.2} & 27.3 $\pm$ 0.2 & \textbf{67.7 $\pm$ 0.3 } & 37.0 $\pm$ 0.3 & \textbf{30.3 $\pm$ 0.2} & 29.3 $\pm$ 0.3 & \textbf{34.5 $\pm$ 0.3} & 67.6 $\pm$ 0.3  \\
     BiT-ResNet-101-3& 12.7 $\pm$ 0.2 & 24.5 $\pm$ 0.2 & 19.7 $\pm$ 0.4 & 34.4 $\pm$ 0.3 & 15.9 $\pm$ 0.2 & \textbf{30.7 $\pm$ 0.3} & 13.5 $\pm$ 0.2 & 91.9 $\pm$ 0.2 \\
     BiT-ResNet-152-4& 12.4 $\pm$ 0.2 & 26.9 $\pm$ 0.2 & 30.8 $\pm$ 0.5 & 35.8 $\pm$ 0.3 & 16.0 $\pm$ 0.2 & \textbf{30.7 $\pm$ 0.3} & 18.7 $\pm$ 0.3 &  91.4 $\pm$ 0.2 \\
     BiT-ResNet-50-1& 15.8 $\pm$ 0.2 & \textbf{27.5 $\pm$ 0.2} & 47.6 $\pm$ 0.4 & \textbf{37.4 $\pm$ 0.3} & 19.7 $\pm$ 0.2 & 30.6 $\pm$ 0.3 & 21.7 $\pm$ 0.2 & \textbf{93.4 $\pm$ 0.2} \\
\bottomrule

\end{tabular}
\end{small}
\vspace{5 pt}
\caption{Comparing a few-shot learner utilizing the full library of nine ILSVRC2012-trained deep CNNs with the larger CNNs trained on the full ImageNet.}
\label{tab:vs_bigt_1shot}
\end{table*}

\begin{table*}[t!]
\centering
\begin{small}

\begin{tabular}{l|cccccccr} 
\toprule
     \multicolumn{1}{c}{} & Aircraft & FC100 & Omniglot & Texture & Traffic & Fungi & Quick Draw & \multicolumn{1}{c}{VGG Flower} \\ 
\midrule 
      \multicolumn{1}{c}{} & \multicolumn{8}{c}{5-way, 1-shot} \\
\midrule 
     Full Library & 44.9 $\pm$ 0.9 & 60.9 $\pm$ 0.9 & \textbf{88.4 $\pm$ 0.7} & \textbf{68.4 $\pm$ 0.8} & 66.2 $\pm$ 0.9 & 57.7 $\pm$ 1.0 & \textbf{65.4 $\pm$ 0.9} & \textbf{86.3 $\pm$ 0.8}\\
     Hard Ensemble & 45.0 $\pm$ 0.9 & 60.0 $\pm$ 0.9 & \textbf{88.4 $\pm$ 0.6} & 67.9 $\pm$ 0.9 & 65.2 $\pm$ 0.8 & \textbf{58.1 $\pm$ 0.9}  & 64.7 $\pm$ 1.0 & 84.9 $\pm$ 0.8 \\
     Soft Ensemble & 44.2 $\pm$ 0.9 & 61.0 $\pm$ 0.9 & 88.2 $\pm$ 0.6 & 67.4 $\pm$ 0.9 & 63.2 $\pm$ 0.8 & 57.1 $\pm$ 1.0 & 65.2 $\pm$ 0.9 & \textbf{86.3 $\pm$ 0.7} \\
     Best Single & \textbf{46.1 $\pm$ 1.0} & \textbf{61.2 $\pm$ 0.9} & 86.5 $\pm$ 0.6 & 65.1 $\pm$ 0.9 & \textbf{66.6 $\pm$ 0.9} & 56.6 $\pm$ 0.9 & 62.8 $\pm$ 0.9 & 83.2 $\pm$ 0.8\\
\midrule
     \multicolumn{1}{c}{} & \multicolumn{8}{c}{20-way, 1-shot} \\
\midrule 
     Full Library & \textbf{25.2 $\pm$ 0.3} & \textbf{36.8 $\pm$ 0.4} & \textbf{76.4 $\pm$ 0.4} & \textbf{46.3 $\pm$ 0.4} & \textbf{40.0 $\pm$ 0.4} & \textbf{37.6 $\pm$ 0.5} & \textbf{44.2 $\pm$ 0.5} & \textbf{75.5 $\pm$ 0.5}\\
     Hard Ensemble & 23.9 $\pm$ 0.3 & 36.3 $\pm$ 0.4 & 73.4 $\pm$ 0.4 & 45.7 $\pm$ 0.4 & 38.2 $\pm$ 0.4 & 36.4 $\pm$ 0.4  & 42.7 $\pm$ 0.4 & 73.3 $\pm$ 0.5 \\
     Soft Ensemble & 24.2 $\pm$ 0.3 & 36.4 $\pm$ 0.4 & 73.8 $\pm$ 0.4 & \textbf{46.3 $\pm$ 0.5} & 37.7 $\pm$ 0.4 & 37.1 $\pm$ 0.4 & 43.4 $\pm$ 0.5 & 74.3 $\pm$ 0.5 \\
     Best Single & 24.3 $\pm$ 0.3 & 36.4 $\pm$ 0.4 & 69.1 $\pm$ 0.4 & 42.5 $\pm$ 0.4 & 38.5 $\pm$ 0.4 & 33.9 $\pm$ 0.5 & 39.5 $\pm$ 0.5 & 70.0 $\pm$ 0.5\\
\midrule
     \multicolumn{1}{c}{} & \multicolumn{8}{c}{40-way, 1-shot} \\
\midrule 
     Full Library & \textbf{18.6 $\pm$ 0.2} & 27.3 $\pm$ 0.2 & \textbf{67.7 $\pm$ 0.3 } & \textbf{37.0 $\pm$ 0.3} & \textbf{30.3 $\pm$ 0.2} & \textbf{29.3 $\pm$ 0.3} & \textbf{34.5 $\pm$ 0.3} & \textbf{67.6 $\pm$ 0.3} \\
     Hard Ensemble& 17.7 $\pm$ 0.2 & 27.3 $\pm$ 0.2 & 65.8 $\pm$ 0.3 & 36.3  $\pm$ 0.3 & 29.0 $\pm$ 0.2 & 28.7 $\pm$ 0.3 & 33.7 $\pm$ 0.3 & 66.1 $\pm$ 0.3  \\
     Soft Ensemble & 18.1 $\pm$ 0.2 & \textbf{27.6 $\pm$ 0.3} & 66.2 $\pm$ 0.3 & \textbf{37.0 $\pm$ 0.3} & 29.0 $\pm$ 0.2 & 29.1 $\pm$ 0.2 & 34.1 $\pm$ 0.3 & 67.4 $\pm$ 0.3 \\
     Best Single & 17.4 $\pm$ 0.2 & 27.2 $\pm$ 0.3 & 61.6 $\pm$ 0.3 & 33.2 $\pm$ 0.3 & 29.5 $\pm$ 0.2 & 26.8 $\pm$ 0.3 & 31.2 $\pm$ 0.3 & 62.8 $\pm$ 0.3 \\
\bottomrule

\end{tabular}
\end{small}
\vspace{5 pt}
\caption{Accuracy obtained using all nine library CNNs as the basis for a few-shot learner.}
\label{tab:ensemble_1shot}
\end{table*}

\begin{table*}[t!]
\centering
\begin{small}

\begin{tabular} {l|cccr}

\toprule
     \multicolumn{1}{c}{} & {Number of} & Hidden & Learning & \multicolumn{1}{c}{Regularization}\\
      \multicolumn{1}{c}{} & Epochs & Size & Rate & Constant \\
\midrule
     \multicolumn{1}{c}{} & \multicolumn{4}{c}{5-way, 1-shot and 5-shot} \\
\midrule
     DenseNet121 & 200 & 1024 & $1 \times 10^{-3}$ & 0.2 \\
     DenseNet161 & 100 & 1024 & $5 \times 10^{-4}$ & 0.2 \\
     DenseNet169 & 300 & 1024 & $5 \times 10^{-4}$ & 0.5 \\
     DenseNet201 & 100 & 512 & $5 \times 10^{-4}$ & 0.5 \\
     ResNet18 & 200 & 512 & $1 \times 10^{-3}$ & 0.2 \\
     ResNet34 & 100 & 1024 & $5 \times 10^{-4}$ & 0.2 \\
     ResNet50 & 300 & 2048 & $5 \times 10^{-4}$ & 0.1 \\
     ResNet101 & 100 & 512 & $1 \times 10^{-3}$ & 0.1 \\
     ResNet152 & 300 & 512 & $5 \times 10^{-4}$ & 0.1 \\
     Full Library & 300 & 1024 & $5 \times 10^{-4}$ & 0.1 \\
     BiT-ResNet-101-3 & 300 & 4096 & $1 \times 10^{-3}$ & 0.7 \\
     BiT-ResNet-152-4 & 300 & 2048 & $5 \times 10^{-4}$ & 0.7 \\
     BiT-ResNet-50-1 & 200 & 2048 & $5 \times 10^{-4}$ & 0.5 \\
\midrule
     \multicolumn{1}{c}{} & \multicolumn{4}{c}{20-way, 1-shot and 5-shot} \\
\midrule
     DenseNet121 & 100 & 1024 & $5 \times 10^{-4}$ & 0.2 \\
     DenseNet161 & 100 & 512 & $1 \times 10^{-3}$ & 0.1 \\
     DenseNet169 & 300 & 512 & $5 \times 10^{-4}$ & 0.1 \\
     DenseNet201 & 200 & 1024 & $5 \times 10^{-4}$ & 0.1 \\
     ResNet18 & 200 & 2048 & $5 \times 10^{-4}$ & 0.1 \\
     ResNet34 & 100 & 1024 & $5 \times 10^{-4}$ & 0.1 \\
     ResNet50 & 100 & 1024 & $5 \times 10^{-4}$ & 0.1 \\
     ResNet101 & 200 & 2048 & $5 \times 10^{-4}$ & 0.2 \\
     ResNet152 & 100 & 512 & $5 \times 10^{-4}$ & 0.2 \\
     Full Library & 100 & 512 & $5 \times 10^{-4}$ & 0.1 \\
     BiT-ResNet-101-3 & 300 & 2048 & $5 \times 10^{-4}$ & 0.5 \\
     BiT-ResNet-152-4 & 300 & 1024 & $5 \times 10^{-4}$ & 0.5 \\
     BiT-ResNet-50-1 & 100 & 2048 & $5 \times 10^{-4}$ & 0.9 \\
\midrule
     \multicolumn{1}{c}{} & \multicolumn{4}{c}{40-way, 1-shot and 5-shot} \\
\midrule
     DenseNet121 & 100 & 2048 & $5 \times 10^{-4}$ & 0.1 \\
     DenseNet161 & 100 & 512 & $5 \times 10^{-4}$ & 0.1 \\
     DenseNet169 & 100 & 512 & $1 \times 10^{-3}$ & 0.2 \\
     DenseNet201 & 100 & 1024 & $5 \times 10^{-4}$ & 0.1 \\
     ResNet18 & 100 & 512 & $1 \times 10^{-3}$ & 0.1 \\
     ResNet34 & 100 & 2048 & $5 \times 10^{-4}$ & 0.2 \\
     ResNet50 & 100 & 512 & $5 \times 10^{-4}$ & 0.1 \\
     ResNet101 & 100 & 512 & $5 \times 10^{-4}$ & 0.1 \\
     ResNet152 & 100 & 1024 & $5 \times 10^{-4}$ & 0.1 \\
     Full Library & 100 & 1024 & $5 \times 10^{-4}$ & 0.1 \\
     BiT-ResNet-101-3 & 300 & 512 & $5 \times 10^{-4}$ & 0.7 \\
     BiT-ResNet-152-4 & 200 & 1024 & $5 \times 10^{-4}$ & 0.5 \\
     BiT-ResNet-50-1 & 300 & 1024 & $5 \times 10^{-4}$ & 0.5 \\
\bottomrule
\end{tabular}
\end{small}
\vspace{5 pt}
\caption{Hyper-parameter settings for different backbones in 5, 20, and 40 ways.}
\label{tab:hyper_parameter}
\end{table*}

\end{document}